\documentclass[10pt,twocolumn,letterpaper]{article}

\usepackage[pagenumbers]{wacv}      

\newcommand{\ABB}[0]{DBAC}

\usepackage{pifont}
\newcommand{\xmark}{\ding{55}}
\newcommand{\cmark}{\ding{51}}
\newcommand*\colourcheck[1]{%
  \expandafter\newcommand\csname #1check\endcsname{\textcolor{#1}{\cmark}}%
}
\newcommand*\colourx[1]{%
  \expandafter\newcommand\csname #1x\endcsname{\textcolor{#1}{\xmark}}%
}
\colourcheck{green}
\colourx{red}
\colourcheck{teal}
\newcommand{\green}[1]{\textcolor{OliveGreen}{\scriptsize #1}}

\usepackage{tabularx}
\usepackage{makecell}
\usepackage{multirow}
\usepackage{comment}
\usepackage{booktabs}
\usepackage{subcaption}
\newcolumntype{Y}{>{\centering\arraybackslash}X}
\newcolumntype{P}[1]{>{\centering\arraybackslash}p{#1}}

\DeclareMathSizes{10}{9}{7}{5}

\definecolor{wacvblue}{rgb}{0.21,0.49,0.74}
\usepackage[pagebackref,breaklinks,colorlinks,allcolors=wacvblue]{hyperref}

\def\wacvPaperID{3071} 
\def\confName{WACV}
\def\confYear{2026}

\title{A Woman with a Knife or A Knife with a Woman? \\ Measuring Directional Bias Amplification in Image Captions}

\author{Rahul Nair *\\
Arizona State University\\
{\tt\small rnair21@asu.edu}
\and
Bhanu Tokas *\\
Arizona State University\\
{\tt\small bhanu.tokas@asu.edu}
\and
Hannah Kerner\\
Arizona State University\\
{\tt\small hkerner@asu.edu}
}

\begin{document}
\maketitle
\begin{abstract}

When we train models on biased datasets, they not only reproduce data biases, but can worsen them at test time --- a phenomenon called bias amplification. Many of the current bias amplification metrics (e.g., $BA_{\rightarrow}$, DPA) measure bias amplification only in classification datasets. These metrics are ineffective for image captioning datasets, as they cannot capture the language semantics of a caption. Recent work introduced Leakage in Captioning (LIC), a language-aware bias amplification metric that understands caption semantics. However, LIC has a crucial limitation: it cannot identify the source of bias amplification in captioning models. We propose Directional Bias Amplification in Captioning (DBAC), a language-aware and directional metric that can identify when captioning models amplify biases. DBAC has two more improvements over LIC: (1) it is less sensitive to sentence encoders (a hyperparameter in language-aware metrics), and (2) it provides a more accurate estimate of bias amplification in captions. Our experiments on gender and race attributes in the COCO captions dataset show that DBAC is the only reliable metric to measure bias amplification in captions.
\end{abstract}
    
\section{Introduction}
\label{sec:intro}

When we train models on biased datasets, they often learn and reproduce those biases in their predictions. In some cases, models can even amplify these biases, producing outputs that are more biased than the data they were trained on. This phenomenon is known as bias amplification.

In this work, we measure bias amplification in image captioning --- do captioning models generate more biased captions than the human-written captions they were trained on? Captioning models are widely used for automatic image tagging and in applications that support the visually impaired \cite{tiwary2023accurate, zhang2023recognize}. Since these models are trained on human captions, they can amplify the societal biases in these captions (e.g., gender bias) \cite{LIC}. These models may eventually pass the amplified biases into downstream tasks --- for instance, assigning the incorrect gender to a person. In sensitive cases, such as describing an image to a visually impaired person, such amplified biases can result in misleading or harmful descriptions.

Several classification-based metrics ($BA_{MALS}$ \cite{zhao-etal-2017-men}, $BA_{\rightarrow}$ \cite{DBA}, DPA \cite{tokas2024makingbiasamplificationbalanced}) have been proposed to measure bias amplification in classification datasets. As shown in \cite{LIC, Image_Cap_2}, these metrics cannot be used to measure bias amplification in captioning models. This is because classification-based metrics do not understand the semantics of the caption. 

To measure bias amplification between model-generated captions (MGCs) and human-generated captions (HGCs), recent works introduced two specialized metrics: $ImgCap^2$~\cite{Image_Cap_2} and LIC~\cite{LIC}. $ImgCap^2$  entangles the biases between images and captions. With $ImgCap^2$, it is challenging to determine whether the metric reflects bias amplification in the captioning model or biases inherent in the image. LIC is an effective language-aware metric for captions. It uses sentence encoders (e.g., BERT) to measure bias amplification.

LIC has a crucial limitation --- it cannot identify the source of bias amplification in captioning models. Consider a model-generated caption --- a woman is holding a knife. LIC cannot identify if the captioning model amplified the bias towards predicting a woman (gender) when it saw a knife (object or task) in the image, or towards predicting a knife when it saw a woman. 

To conduct targeted bias mitigation, we need a metric that can identify the source of bias amplification. In the previous example, suppose a captioning model amplified bias toward women when it saw a knife. In this case, instead of retraining the captioning model with a much larger dataset, we should retrain it only on images of women without knives to mitigate this bias. 

We need a metric that: $(1)$ is language-aware: understands caption semantics, and $(2)$ is directional: measures the bias amplification caused by attributes on tasks, and by tasks on attributes. We introduce Directional Bias Amplification in Captioning (DBAC), the first language-aware and directional metric for captions.

Our key contributions: (1) With DBAC, users can identify the source of bias amplification in captioning models. Compared to LIC, DBAC offers additional improvements: (2) DBAC scores are consistent across sentence encoders. This eliminates the pain of choosing a suitable encoder to measure bias amplification. (3) DBAC uses contextual substitution for vocabulary alignment, which is better than constant substitution used in LIC (vocabulary alignment is an important step in language-aware metrics). With contextual substitution, both LIC and DBAC report more accurate bias amplification scores. 

We performed extensive experiments using gender and race attributes in the COCO captions dataset \cite{chen2015microsoft}. We showed that DBAC is the only metric that can reliably measure bias amplification in captioning models.

\section{Desirable Properties in Bias Amplification Metrics for Image Captioning}

Our goal is to design a metric to measure bias amplification in captioning models. We draw on literature~\cite{LIC, DBA, Image_Cap_2, tokas2024makingbiasamplificationbalanced} to identify the desired properties in such a metric.


\begin{itemize}
    \item \textbf{Language Awareness:} Several bias amplification metrics (e.g., $BA_{MALS}$, $BA_{\rightarrow}$, DPA) only use attribute $A$ (e.g., gender, race) and task $T$ (e.g., object, activity) words to quantify bias in a caption. This approach overlooks caption semantics. For instance, the captions ``He has a ball" and ``He does not have a ball" have the same gender (He) and task (ball). However, they have different meanings and are not equivalent in terms of bias.

    \item \textbf{Directionality:} To inform bias mitigation efforts, a metric should not only report the magnitude of bias, but also its direction. For example, if a captioning model amplified the bias toward predicting a knife, $T$, when it saw a woman, $A$ ($A \rightarrow T$), we should train the model with more images of women with non-knife objects to mitigate the bias. If the model amplified bias towards predicting a woman when it saw a knife ($T \rightarrow A$), we should train it with more images of men with a knife to mitigate the bias.

    \item \textbf{Consistency:} If a metric has multiple hyperparameters, the bias amplification score should be similar for different parameter choices. Otherwise, users cannot determine when the metric reflects reality. With an inconsistent metric, the same model will appear highly biased under one set of hyperparameters but unbiased under another.
    
\end{itemize}

\section{Related Works}
\label{sec:rel_work}

\paragraph{Bias Amplification} 
\citet{zhao-etal-2017-men} proposed the first bias amplification metric, $BA_{MALS}$, for machine learning models, specifically for multi-label object classification and visual semantic role labeling models. $BA_{MALS}$ measured increases in co-occurrences of the majority $A-T$ pairs from the dataset. \citet{intersectional_defintion_of_fairness} measured bias amplification using the difference in ``differential fairness'', a measure of the difference in co-occurrences of $A-T$ pairs across different values of $A$.

Subsequent work proposed application-specific measures. \citet{lin2019crankvolumepreferencebias} proposed a new metric called bias disparity to measure bias amplification in recommender systems. \citet{seshadri-etal-2024-bias} measured bias amplification for text-to-image generation using the increase in percentage bias in generated samples vs. training samples.

\vspace{-0.2cm}
\paragraph{Bias Amplification for Image Captioning}
\citet{Hendricks_2018_ECCV} was among the first to specifically measure bias in image captioning. They introduced two metrics: $Gender\ error$, which calculates the number of incorrectly predicted gender words, and $Gender\ ratio$, which classifies each caption as male or female based on the presence of some predefined gender-indicating word, then computes the ratio of male captions to female captions.

\citet{LIC} later proposed the metric, LIC, for measuring bias amplification in captioning. Unlike existing metrics, LIC is language-aware. \citet{Image_Cap_2} proposed another metric, $ImgCap^2$, which used prompt-based image captioning models to measure bias amplification. It assumes masked images to have no bias, which was disproven by \citet{meister2023gender}. $ImgCap^2$ confounds the biases from the masked images and the captions. We want to measure the increase in biases only in the captions. $ImgCap^2$ is unsuitable as it measures a different quantity. \citet{Image_Cap_2} showed that the LIC metric can have inconsistent bias amplification scores based on the choice of sentence encoders.

\begin{table}[htbp]
    \vspace{-0.2cm}
    \begin{center}        
    \footnotesize
    \begin{tabular}{cccc}
        \toprule
         \textbf{Method} & \makecell{\textbf{Language} \\\textbf{Awareness}} & \textbf{Directionality} & \textbf{Consistency}\\
         \hline
         $ImgCap^2$~\cite{Image_Cap_2} & \tealcheck & \redx & \tealcheck\\
         LIC~\cite{LIC} & \tealcheck & \redx & \redx\\
         $BA_{\rightarrow}$~\cite{DBA} & \redx & \tealcheck & \tealcheck\\
         DPA~\cite{tokas2024makingbiasamplificationbalanced} & \redx & \tealcheck & \tealcheck \\
         \ABB\ (Ours) & \tealcheck & \tealcheck & \tealcheck\\
         \bottomrule
    \end{tabular}
    \end{center}
    \vspace{-0.4cm}
    \caption{Comparison of various bias amplification metrics. Only DBAC satisfies all three desirable properties of a bias amplification metric for image captioning.}
    \vspace{-0.5cm}
    \label{tab:metric_comparision}
\end{table}

\vspace{-0.2cm}
\paragraph{Directionality in Bias Amplification}
All of the metrics discussed in the previous paragraphs can measure the magnitude of bias amplification. None of these can measure the direction in which biases are amplified. \citet{DBA} introduced the first metric, $BA_{\rightarrow}$, capable of disentangling the direction of bias amplification for classification tasks. \citet{MDBA} created $Multi_{\rightarrow}$, a new metric capable of measuring directional bias amplification in multiple attributes. However, $Multi_{\rightarrow}$ cannot differentiate between positive and negative bias amplification (i.e., increase or decrease in biases). Recently, \citet{tokas2024makingbiasamplificationbalanced} introduced the DPA metric, which measures directional bias amplification using the concept of predictability. However, $BA_{\rightarrow}$, $Multi_{\rightarrow}$, and DPA are not language-aware, making them unsuitable for captioning. 

We compared bias amplification metrics in Table \ref{tab:metric_comparision}.
\section{Methodology} \label{subsec: Formulation}

\subsection{Proposed Metric}
In image captioning, each image is annotated with the gender of the person: $a$ (male, female), and the objects (or tasks) associated with them: $t$ (e.g., bat, ball). We ask a human and a model to generate a caption describing each image. Bias amplification is defined as the difference in bias between the model-generated captions (MGCs) and human-generated captions (HGCs). We discuss DBAC, our metric to measure bias amplification caused by gender on tasks ($A \rightarrow T$), and by tasks on gender ($T \rightarrow A$).

We first discuss how to measure $A \rightarrow T$ bias amplification using DBAC. Consider an HGC, which we denote as $s$: ``A man with a guitar''. We masked gender-revealing tokens in $s$. ``A man with a guitar'' becomes ``A $<$gender$>$ with a guitar''. We denote the gender-masked HGC as $s_{a}$. We performed an additional processing step on $s_{a}$ called vocabulary alignment, which we discussed in Section \ref{subsec:vocab_align}.

We trained a model $f^{a}_{s_{a}}$ to classify the gender ($a$) of the person in the image using the gender-masked HGC ($s_{a}$) as an input. We call this model the attacker function. Our attacker function is a combination of a sentence encoder (e.g., LSTM) and a classification head (e.g., MLP). We measured the performance of our attacker using classification accuracy, which we denote as $Q$ \footnote{$Q$ should be a positive-valued function. The lower bound of $Q$ should indicate the attacker function's worst performance. The upper bound of $Q$ should indicate the attacker function's best performance.}. To measure $Q$, we used the gender label ($a$) from the image as ground-truth. 


If the attacker function $f^{a}_{s_{a}}$ can predict gender ($a$) with high accuracy using $s_{a}$, it shows that \emph{$a$} can be inferred from the HGC $s_{a}$, without explicit gender information. This indicates a bias in the HGCs. We denote this bias as $\omega_{H: A \rightarrow T}$. In probabilistic terms, $\omega_{H: A \rightarrow T}$ is proportional to $P(a|s_{a})$.
\vspace{-0.1cm}
\begin{equation} \label{eq:hgc_bias_t_a}    
    \omega_{H:A \rightarrow T} =  Q(f^{a}_{s_{a}}(s_{a}),a) \approx P(a|s_{a})
    \end{equation} 
\vspace{-0.1cm}
For the above HGC, consider the corresponding MGC: ``A boy with a guitar". We denote this caption as $\hat{s}$. We masked gender-revealing tokens in $\hat{s}$ to get the gender-masked MGC $\hat{s}_{a}$. We trained an attacker function $f^{a}_{\hat{s}_{a}}$ to predict \emph{$a$} using $\hat{s}_{a}$ as an input. The performance of the attacker function $f^{a}_{\hat{s}_{a}}$ indicates a bias in the MGC, which we denote as $\omega_{M: A  \rightarrow T}$. In probabilistic terms, $\omega_{M:A \rightarrow T}$ is proportional to $P(a|\hat{s}_{a})$.
\vspace{-0.2cm}
\begin{equation} \label{eq:mgc_bias_t_a}    
    \omega_{M:A \rightarrow T} =  Q(f^{a}_{\hat{s}_{a}}(\hat{s}_{a}),a) \approx P(a|\hat{s}_{a})
    \end{equation} 

To measure bias amplification, we typically measure the change in biases between the MGC and HGC: $\omega_{M: A \rightarrow T} - \omega_{H: A \rightarrow T}$. This is proportional to $P(a|\hat{s}_{a})$ $-$ $P(a|s_{a})$. Here, we have two different priors: $\hat{s}_{a}$ and $s_{a}$. According to \cite{DBA}, bias amplification (or difference in bias) should be measured with respect to a fixed prior variable. 

To ensure a fixed prior, we made $\omega_{H:A \rightarrow T}$ proportional to $P(s_{a}|a)$, instead of $P(a|s_{a})$. We made $\omega_{M:A \rightarrow T}$ proportional to $P(\hat{s}_{a}|a)$, instead of $P(a|\hat{s}_{a})$. We used the Bayes' theorem, as shown in equations \ref{eq:hgc_bias_t_a_bayes} and \ref{eq:mgc_bias_t_a_bayes}. In equation \ref{eq:hgc_bias_t_a_bayes}, $P(a)$ refers to the probability of the ground-truth gender in the image. $P(s_{a})$ refers to the probability of the sentence $s_{a}$ (A $<$gender$>$ with a guitar) occuring in the HGCs. 

\vspace{-0.2cm}
\begin{equation} \label{eq:hgc_bias_t_a_bayes}    
    \omega_{H:A \rightarrow T} =  Q(f^{a}_{s_{a}}(s_{a}),a) \cdot \frac{P(s_{a})}{P(a)} \approx P(s_{a}|a)
\end{equation}
\vspace{-0.2cm}
\begin{equation} \label{eq:mgc_bias_t_a_bayes}    
    \omega_{M:A \rightarrow T} =  Q(f^{a}_{\hat{s}_{a}}(\hat{s}_{a}),a) \cdot \frac{P(\hat{s}_{a})}{P(a)} \approx P(\hat{s}_{a}|a)
\end{equation}

We cannot compute $P(s_{a})$, as the probability of a sentence is ill-defined. In $s_{a}$, we assumed that the only relevant information is the task $t$ (guitar). We approximated $P(s_{a})$ $\approx$ $P(t)$ in equation \ref{eq:hgc_bias_t_a_bayes} to get equation \ref{eq:hgc_bias_t_a_bayes_approximated}. $P(t)$ refers to the probability of the word \emph{guitar} occurring across all HGCs. As $P(s_{a})$ $\approx$ $P(t)$, $\omega_{H:A \rightarrow T}$ can be better approximated as $P(t|a)$. We performed a similar approximation for the MGC $\hat{s}_{a}$: $P(\hat{s}_{a})$ $\approx$ $P(\hat{t})$, to get equation \ref{eq:mgc_bias_t_a_bayes_approximated}. $P(\hat{t})$ refers to the probability of \emph{guitar} occurring across all MGCs. 



\begin{equation} \label{eq:hgc_bias_t_a_bayes_approximated}    
    \omega_{H:A \rightarrow T} =  Q(f^{a}_{s_{a}}(s_{a}),a) \cdot \frac{P(t)}{P(a)} \approx P(t|a)
    \end{equation} 

\vspace{-0.2cm}

\begin{equation} \label{eq:mgc_bias_t_a_bayes_approximated}    
    \omega_{M:A \rightarrow T} =  Q(f^{a}_{\hat{s}_{a}}(\hat{s}_{a}),a) \cdot \frac{P(\hat{t})}{P(a)} \approx P(\hat{t}|a)
    \end{equation}

To compute bias amplification, we computed $\omega_{M:A \rightarrow T}$ $-$ $\omega_{H:A \rightarrow T}$ using equations \ref{eq:hgc_bias_t_a_bayes_approximated} and \ref{eq:mgc_bias_t_a_bayes_approximated}. The final equation is shown in equation \ref{eq:a_t_bias_amp}. This is proportional to $P(\hat{t}|a)-P(t|a)$. Equation \ref{eq:a_t_bias_amp} for bias amplification follows the definition of directionality as it has a fixed prior variable $a$ \cite{DBA}. Here, we measured how the attribute $a$ amplifies the bias in the tasks: $\hat{t}$ vs $t$. This is called $A \rightarrow T$ bias amplification \cite{DBA}. 

Note: Approximating $P(s_{a})$ by $P(t)$ assumes non-task tokens in the sentence $s_{a}$ contribute negligibly to the attacker. If non-task tokens carry a residual bias signal, DBAC may under-/over-estimate amplification.

In equation \ref{eq:a_t_bias_amp}, we normalized DBAC to ensure bias amplification is in a fixed range of $(-1, 1)$. We added an $\epsilon$ term ($\epsilon \to 0^{+}$) to the denominator for numerical stability. As shown in \cite{tokas2024makingbiasamplificationbalanced}, normalizing a bias amplification metric makes it robust to attacker functions. A DBAC$_{A \rightarrow T}$ score close to $-1$ indicates that the captioning model (MGCs) significantly mitigated $A \rightarrow T$ bias in the HGCs. A score close to $1$ indicates that the captioning model significantly worsened the $A \rightarrow T$ bias. 
\vspace{-0.2cm}
\begin{equation} \label{eq:a_t_bias_amp}
        DBAC_{A \rightarrow T} = \frac{\omega_{M:A \rightarrow T} - \omega_{H:A \rightarrow T}}{\omega_{M:A \rightarrow T} + \omega_{H:A \rightarrow T} + \epsilon}
    \end{equation}

To measure $T \rightarrow A$ bias amplification, we followed a similar approach. We masked task-revealing tokens in both the HGCs and MGCs. For the HGC $s$: ``A man with a guitar'', the task-masked HGC $s_{t}$ is: ``A man with a $<$Obj$>$''. Our final equation for $T \rightarrow A$ is shown in equation \ref{eq:t_a_bias_amp}. Here, $t$ refers to the ground-truth task labels in the image.
\vspace{-0.1cm} 
\begin{equation} \label{eq:t_a_bias_amp}
        DBAC_{T \rightarrow A} = \frac{\omega_{M:T \rightarrow A} - \omega_{H:T \rightarrow A}}{\omega_{M:T \rightarrow A} + \omega_{H:T \rightarrow A} + \epsilon}
    \end{equation}

\subsection{Vocabulary Alignment}
\label{subsec:vocab_align}

Since HGCs are written by humans, they tend to have a larger vocabulary (number of unique words) than the MGCs. This difference in vocabulary can underestimate bias amplification scores \cite{LIC}. To ensure that HGCs and MGCs use a similar vocabulary, we performed a processing step called vocabulary alignment, similar to LIC. 

Let us call the vocabulary of HGCs (the number of unique words) $V_{HGC}$ and the vocabulary of MGCs $V_{MGC}$. We replaced extra words in $V_{HGC}$ (that were not present in $V_{MGC}$) with their closest matching word in $V_{MGC}$. Equation \ref{eq: Vocab Sub} summarizes our vocabulary alignment technique. Consider a word $v$ (e.g., seat) from $V_{HGC}$. If $v$ is not present in $V_{MGC}$, we calculated the semantic distance between $v$ and each word in $V_{MGC}$. This distance is indicated by $dist(v, v')$. To compute $dist(v, v')$, we can use any word embedding model. For our experiments, we tested FastText \cite{bojanowski2016enriching} and GloVe \cite{pennington-etal-2014-glove}.

We identify the word $v'$ that has the minimum semantic distance from $v$. If $v'$ (e.g., chair) falls within a user-defined threshold $\delta$, we replace all occurrences of $v$ in the HGCs with $v'$ (replace all occurrences of ``seat'' with ``chair''). Else, we replace all occurrences of $v$ with $<$unk$>$. We conducted a sensitivity analysis of $\delta$ values in Section \ref{sec: glove thresholds}.
\vspace{-0.2cm}
\begin{equation} \label{eq: Vocab Sub}
    v_{new} = 
    \begin{cases} 
        v'\ s.t. \min\limits_{v' \in V_{MGC}} dist(v,v'), & dist(v,v') < \delta \\
        <\text{unk}>, & otherwise   
    \end{cases}
\end{equation}
\noindent

LIC replaces all occurrences of $v$ in the HGCs with $<$unk$>$. We refer to LIC's substitution technique as constant substitution, and we call our technique contextual substitution. In Section \ref{subsec:constant_vs_contextual}, we showed that contextual substitution provides a more accurate estimate of bias amplification than constant substitution.

\section{Experiments and Results}
We conducted our experiments on the COCO captions dataset \cite{chen2015microsoft}. We used the gender (male, female) and race (light skin, dark skin, unsure) annotations from COCO captions. For the $10780$ images, we used an $80$-$20$ data split, resulting in $8624$ training and $2156$ test images. For experiment \ref{subsec:lang_based_bias_exp}, we evaluated the gender attribute. For all other experiments, we evaluated gender and race attributes.

We measured bias amplification in $13$ captioning models: FC \cite{rennie2017self}, Att2In \cite{rennie2017self}, UpDown \cite{UpDn}, Oscar \cite{li2020oscar}, NIC+Equalizer \cite{Hendricks_2018_ECCV}, SAT \cite{xu2015show}, NIC \cite{vinyals2015show}, Transformer \cite{Attention_is_all_you_need}, BakLLAVA \cite{liu2023improvedllava}, BLIP \cite{li2022blip}, Florence \cite{florence}, LLAVA \cite{liu2023llava}, and ViT\_GPT2 \cite{vit_gpt2}. 

We evaluated two general directional metrics: $BA_{\rightarrow}$ and DPA, and two language-aware metrics:  LIC and DBAC (ours). Additional setup details are in Section \ref{sec:additional_details}.

BakLLAVA, LLAVA, and Florence were not trained on COCO's HGCs. For these models, the scores reported by our chosen metrics cannot be termed bias amplification (as the training dataset is different). These scores indicate a difference in bias between the MGCs and COCO's HGCs. 

We did not study captioning datasets like PAO-evalbias \cite{qiu2023gender} and ArtEmis (V1 and V2) \cite{achlioptas2021artemis, mohamed2022okay}. In PAO-evalbias, the HGCs do not reflect real-world human captions. All images that share the same attribute and task have the same template caption: a \{gender\} with a \{task\}. ArtEmis does not have annotations for attributes and tasks in the image. Without them, we cannot measure bias amplification.

\subsection{Capturing language biases}
\label{subsec:lang_based_bias_exp}

\begin{table}[t]
    \centering
    \footnotesize
    \begin{tabular}{cccc}
        \toprule
        \textbf{Subset} & \textbf{Task} & \textbf{Base Verb1} & \textbf{Base Verb 2} \\
        \midrule
        Subset 1 & bed & lay & sit \\
        Subset 2 & frisbee & throw & play \\
        Subset 3 & umbrella & hold & walk \\
        \bottomrule
    \end{tabular}
    \vspace{-0.2cm}
    \caption{Subset combinations: The table shows the base verbs and the corresponding task in each subset. }
    \label{tab: Language Context Content}
    \vspace{-0.3cm}
\end{table}

\begin{table}[t]
    \centering
    \footnotesize
    \begin{tabular}{P{0.65cm}P{0.85cm}P{1.1cm}P{1.25cm}P{1.05cm}P{0.9cm}}
        \toprule
        \textbf{Dataset} & \textbf{Ratio} &\textbf{$\text{BA}_{H:A \rightarrow T}$} & \textbf{DPA$_{H:A \rightarrow T}$} & $\boldsymbol{\omega}_{H:A \rightarrow T}$  & $\textbf{LIC}_{\textbf{H}}$\\
        \midrule
        $C_{1}$ & $30:20$ & $0.50_{0.00}$ & $0.50_{0.00}$ & $0.64_{0.02}$ & $0.65_{0.01}$\\
        $C_{2}$ & $35:15$ & $0.50_{0.00}$ & $0.50_{0.01}$ & $0.73_{0.03}$ & $0.60_{0.02}$\\
        $C_{3}$ & $40:10$ & $0.50_{0.00}$ & $0.50_{0.00}$ & $0.80_{0.02}$ & $0.60_{0.02}$\\
        $C_{4}$ & $45:05$ & $0.50_{0.00}$ & $0.50_{0.00}$ & $0.89_{0.04}$ & $0.72_{0.03}$\\
        \bottomrule
    \end{tabular}
    \vspace{-0.2cm}
    \caption{Capturing language bias in COCO captions: Across the controlled datasets, $BA_{H:A \rightarrow T}$ and DPA$_{H:A \rightarrow T}$ reported no bias in the HGCs ($0.5$). DBAC ($\omega_{H: A \rightarrow T}$) and LIC$_{H}$ reported positive bias ($> 0.5$). Subscripts show 95\% confidence intervals. } 
    \label{tab: Language Context}
    \vspace{-0.6cm}
\end{table}

\begin{table*}[tbp]
    \footnotesize
    \begin{tabularx}{0.98\textwidth}{P{1cm}P{2.5cm}cYYY}
        \toprule
        \textbf{Attribute} & \textbf{Metric} & \textbf{Original} & \textbf{Partial Masked} & \textbf{Segment Masked} & \textbf{Bounding-Box Masked}\\
        \midrule
         & Image & 
        \includegraphics[width=2.75cm, height=1.7cm]{figures/original.png} & \includegraphics[width=2.75cm, height=1.8cm]{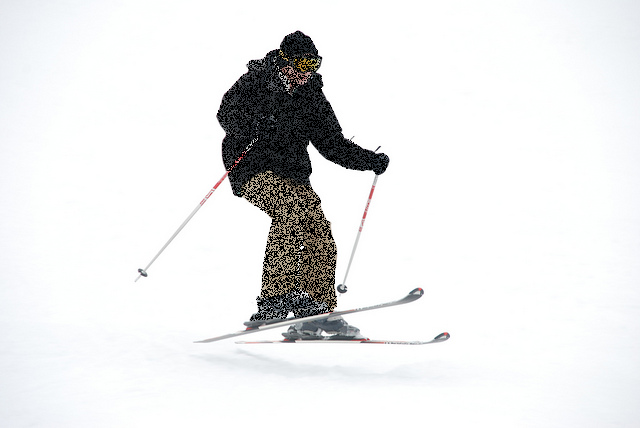} & \includegraphics[width=2.75cm, height=1.7cm]{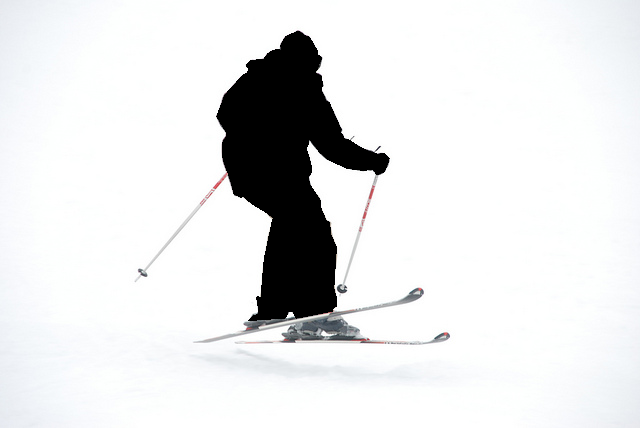} & \includegraphics[width=2.75cm, height=1.7cm]{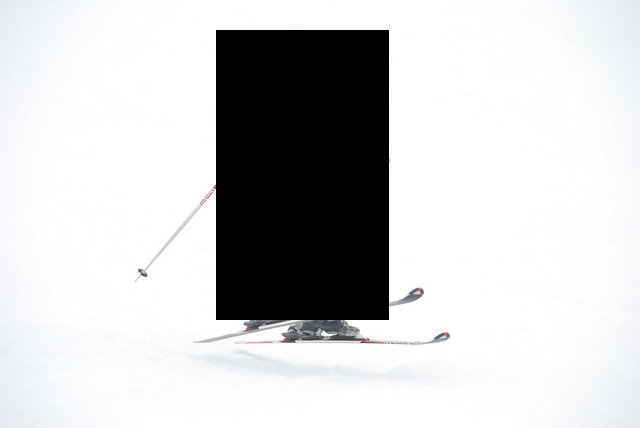}\\
         & Attribution Map &
        \includegraphics[width=2.75cm, height=1.7cm]{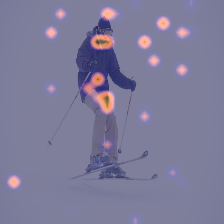} & \includegraphics[width=2.75cm, height=1.7cm]{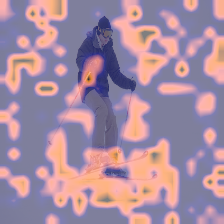} & \includegraphics[width=2.75cm, height=1.7cm]{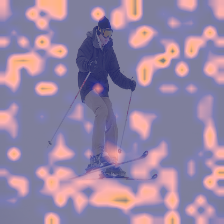} & \includegraphics[width=2.75cm, height=1.7cm]{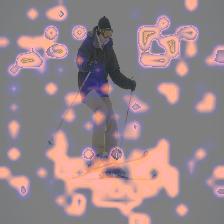}\\
        \midrule
        \multirow{5}{*}{Gender} & Attribution Score & $66.82\%$ & $69.82\%$ & $72.20\%$ & $74.75\%$ \\
         & $\text{LIC}$ & \ \ \ $11.26_{0.92}$ & \ \ \  $08.95_{0.65}$  &  \ \ \ $10.78_{0.82}$ & \ \ \  $00.14_{1.20}$ \\
         & $\text{DPA}_{T\rightarrow A}$ & \ \ \  $00.30_{0.05}$ &  \ \ \ $00.56_{0.07}$ &  \ \ \ $01.12_{0.05}$ &  \ \ \ $01.63_{0.06}$ \\
         & $BA_{T\rightarrow A}$ & \ \ \  $00.41_{0.00}$ &  \ \ \ $00.54_{0.00}$ & \ \ \  $00.83_{0.00}$ & \ \ \  $01.08_{0.00}$ \\
         & $\text{DBAC}_{T\rightarrow A}$ (ours) & \ \ \  $09.98_{0.33}$ &  \ \ \ $17.32_{0.24}$ &  \ \ \ $28.75_{0.31}$ &  \ \ \ $33.42_{0.29}$ \\
        \midrule
        \multirow{5}{*}{Race} & Attribution Score & $71.15\%$ & $74.92\%$ & $75.48\%$ & $79.55\%$ \\
         & $\text{LIC}$ &  \ \ \ $01.11_{0.23}$ & $-11.79_{1.01}$ & \ \ \  $02.13_{0.23}$ & $-05.28_{0.71}$ \\
         & $\text{DPA}_{T\rightarrow A}$ & $-00.22_{0.36}$ &  \ \ \ $10.06_{0.18}$ & $-19.41_{0.21}$ & $-46.60_{0.41}$ \\
         & $BA_{T\rightarrow A}$ & $-00.96_{0.00}$ & $-00.96_{0.00}$ & $-00.61_{0.00}$ & \ \ \  $00.74_{0.00}$ \\
         & $\text{DBAC}_{T\rightarrow A}$ (ours) & \ \ \  $15.57_{0.60}$ & \ \ \  $17.93_{0.53}$ &  \ \ \ $23.39_{0.71}$ &  \ \ \ $31.16_{0.79}$ \\
        \bottomrule
    \end{tabularx}
    \vspace{-0.2cm}
    \caption{Directionality Results: $T \rightarrow A$ bias amplification in BLIP across different masking scenarios. As masking increases, BLIP relies more on background objects to predict the attribute, leading to a steady rise in $T \rightarrow A$ for both gender and race. This increase in $T \rightarrow A$ is captured only by DBAC. All metric values are scaled by $100$. Subscripts show 95\% confidence intervals.}
    \vspace{-0.6cm}
    \label{tab:BLIP Directionality Results}
\end{table*}

\subsubsection{Experiment Setup}
We isolated language bias in COCO's HGCs through a controlled experiment. We examined which metrics can capture this language bias. 

We extracted three subsets from COCO captions. In each subset, we balanced gender and task attributes. For instance, in subset $1$, we included $50$ captions containing a male pronoun (e.g., he, him) and the task ``bed'', and $50$ captions containing a female pronoun (e.g., she, her) and the same task ``bed''. However, we imbalanced the verb words in these captions. Among the male captions, $30$ captions contained lay-based words (e.g., laying, laid) and $20$ captions contained sit-based verbs (e.g., sat, sitting). For the female captions, we reversed the distribution ($20$ lay-based words, and $30$ sit-based words). 

We followed the same process for the other two subsets (keeping a $30$:$20$ verb imbalance). For each subset, the details of the tasks and verb words are provided in Table \ref{tab: Language Context Content}. We combined all three subsets to form our controlled dataset: $C_{1}$. To vary the strength of language bias, we created three more controlled datasets: $C_{2}$ with an $35$:$15$ verb imbalance, $C_{3}$ with a $40$:$10$ verb imbalance, and $C_{4}$ with an $45$:$05$ verb imbalance. We extracted verb annotations manually since COCO captions did not have them.

In all controlled datasets, the gender and task attributes add no bias as they are balanced with each other. The bias arises entirely from language variations in the captions. 

In this experiment, we measured the bias in the HGCs (and not bias amplification). For the directional metrics (DBAC, DPA, $BA_{\rightarrow}$), we measured $A \rightarrow T$ bias.

\subsubsection{Results}
\textbf{Only DBAC and LIC captured language biases in all controlled datasets}. In Table \ref{tab: Language Context}, across the controlled datasets, both $BA_{H: A\rightarrow T}$ and DPA$_{H:A \rightarrow T}$ reported no bias in the HGCs (no bias is indicated by $0.5$). Language-aware metrics like DBAC ($\omega_{H:A \rightarrow T}$) and LIC$_{H}$ correctly reported positive bias ($>0.5$) for all controlled datasets. Metrics like $BA_{\rightarrow}$ and DPA should not be used for captions, as they cannot capture language biases.

\subsection{Directionality}
\label{subsec:directionality}

\subsubsection{Experiment Setup}
We examined which metrics can capture directional bias amplification in captions. We conducted an image masking experiment, similar to \citet{DBA}. 

We modified $T \rightarrow A$ biases in COCO captions by progressively masking people. We created four versions of the dataset: (1) the original dataset and three perturbed versions, where the people in the foreground of the image were progressively masked by: (2) partially masking the segment, (3) completely masking the segment, and (4) completely masking the bounding box. We masked only the people in the foreground, as COCO captions provided gender and race annotations only for them.  

For each dataset, we used BLIP~\cite{li2022blip}, an image captioning model, to generate MGCs. As the masking increases, BLIP's  MGCs will increasingly use task objects to predict gender and race. Since we do not modify the HGCs, the $T\rightarrow A$ amplification in BLIP will gradually increase for both attributes. We evaluated which metrics can capture this increase in $T \rightarrow A$.

\subsubsection{Results}

We measured $T \rightarrow A$ bias in BLIP using GradCam \cite{GradCam}. We calculated an attribution score, which indicates the percentage of non-person pixels (i.e., task objects) that were involved in BLIP's prediction of the person's attribute (gender or race). A high attribution score means that BLIP over-relied on task objects to predict the attribute, indicating a high $T \rightarrow A$ bias.

\textbf{As masking increased, BLIP showed higher attribution scores for gender and race.} In Table \ref{tab:BLIP Directionality Results}, for both gender and race, BLIP's attribution score steadily increased from the original version to the fully masked version. This confirms that as the person was progressively masked, BLIP relied more on task objects to predict gender and race. This led to a rise in $T \rightarrow A$ bias in both attributes. To illustrate this visually, we generated BLIP's attribution maps (for one of the images) for race in Table \ref{tab:BLIP Directionality Results}.

\textbf{For gender, DBAC and $\mathbf{BA_{\rightarrow}}$ captured the growing $\mathbf{T \rightarrow A}$ amplification.} In Table \ref{tab:BLIP Directionality Results}, for gender, DBAC$_{T \rightarrow A}$, DPA$_{T \rightarrow A}$, and $BA_{T \rightarrow A}$ increased with increasing attribution score. LIC did not capture this increasing trend. 

\textbf{For race, only DBAC captured the growing $\mathbf{T \rightarrow A}$ amplification.} In Table \ref{tab:BLIP Directionality Results}, for race, only DBAC$_{T \rightarrow A}$ increased with increasing attribution scores. $BA_{T \rightarrow A}$, DPA$_{T \rightarrow A}$, and LIC did not capture this increasing trend.

For both gender and race, only DBAC correctly captured $T \rightarrow A$ trends as it is both language-aware and directional. Other metrics failed as they are either language-aware (LIC) or directional (DPA and $BA_{\rightarrow}$).

\begin{table}[htbp]
\vspace{-0.2cm}
\footnotesize
\centering

\begin{subtable}{\linewidth}
\centering
\begin{tabularx}{0.8\textwidth}{cccc}
\toprule
\makecell{\textbf{Captioning} \\ \textbf{Models}} & \makecell{\textbf{CV}\\\textbf{(LIC)}} & \makecell{\textbf{CV}\\\textbf{(DBAC)}} & \makecell{\textbf{$\%$}\\ \textbf{Reduction}}\\
\midrule
Att2In~\cite{rennie2017self}      & $2.23$ & $0.78$ & $65.06\%$ \\
BakLLAVA~\cite{liu2023improvedllava}    & $1.56$ & $0.45$ & $71.17\%$ \\
BLIP~\cite{li2022blip}        & $2.84$ & $0.30$ & $89.51\%$ \\
FC ~\cite{rennie2017self}         &$19.54$ \ \ & $0.64$ & $96.71\%$ \\
Florence~\cite{florence}     & $3.60$ & $0.36$ & $89.98\%$ \\
LLAVA~\cite{liu2023llava}        & $1.84$ & $0.24$ & $86.97\%$ \\
Oscar~\cite{li2020oscar}       &$32.52$ \ \ & $0.20$ & $99.38\%$ \\
SAT~\cite{xu2015show}         & $1.08$ & $0.19$ & $82.66\%$ \\
NIC~\cite{vinyals2015show}         & $1.42$ & $0.23$ & $83.52\%$ \\
NIC+Equal.~\cite{Hendricks_2018_ECCV}  & $5.27$ & $0.31$ & $94.15\%$ \\
Transformer~\cite{Attention_is_all_you_need} & $0.85$ & $0.26$ & $69.19\%$ \\
UpDown~\cite{UpDn}      & $2.47$ & $0.35$ & $84.94\%$ \\
VIT\_GPT2~\cite{vit_gpt2}   & $5.53$ & $0.89$ & $83.99\%$ \\
\midrule
\textbf{Average} & -- & -- & $84.48\%$ \\
\bottomrule
\end{tabularx}
\caption{Sentence encoders trained from scratch}
\label{tab:scratch_attackers}
\end{subtable}

\begin{subtable}{0.8\linewidth}
\centering
\begin{tabularx}{\textwidth}{cccc}
\toprule
\makecell{\textbf{Captioning} \\ \textbf{Models}} & \makecell{\textbf{CV}\\\textbf{(LIC)}} & \makecell{\textbf{CV}\\\textbf{(DBAC)}} & \makecell{\textbf{$\%$}\\ \textbf{Reduction}}\\
\midrule
Att2In      &  $0.96$ & $0.18$ &  $81.47\%$ \\
BakLLAVA    &  $6.72$ & $0.22$ &  $96.77\%$ \\
BLIP        &  $0.81$ & $0.76$ &  $06.77\%$ \\
FC          &  $2.35$ & $0.64$ &  $72.57\%$ \\
Florence    &  $2.75$ & $0.18$ &  $93.45\%$ \\
LLAVA       &  $5.52$ & $0.25$ &  $95.38\%$ \\
Oscar       &  $3.50$ & $0.15$ &  $95.61\%$ \\
SAT         &  $3.85$ & $0.54$ &  $85.93\%$ \\
NIC         &  $2.86$ & $1.08$ &  $62.29\%$ \\
NIC+Equal.  &  $1.18$ & $1.02$ &  $13.88\%$ \\
Transformer &  $5.35$ & $0.14$ &  $97.36\%$ \\
UpDown      &  $1.71$ & $0.16$ &  $90.51\%$ \\
VIT\_GPT2   &  $5.57$ & $0.37$ &  $93.41\%$ \\
\midrule
\textbf{Average} & -- & -- & 69.88\% \\
\bottomrule
\end{tabularx}
\caption{Pre-trained sentence encoders}
\label{tab:pretrained_attackers}
\end{subtable}

\vspace{-0.2cm}
\caption{Sensitivity of DBAC and LIC to different sentence encoders (gender results): CV indicates the coefficient of variation in bias amplification scores across encoders. For the encoders trained from scratch, the CV in DBAC was $84.5\%$ less than LIC. For pre-trained encoders, the CV in DBAC was $69.9\%$ less than LIC.}
\label{tab:attackers_comparison}
\vspace{-0.2cm}
\end{table}

\begin{table}[htbp]
\vspace{-0.2cm}
\footnotesize
\centering
\begin{subtable}[t]{0.38\textwidth}
\centering
\begin{tabularx}{\textwidth}{cccc}
\toprule
\makecell{\textbf{Captioning} \\ \textbf{Models}} & \makecell{\textbf{CV}\\\textbf{(LIC)}} & \makecell{\textbf{CV}\\\textbf{(DBAC)}} & \makecell{\textbf{$\%$}\\ \textbf{Reduction}}\\
\midrule
Att2In~\cite{rennie2017self}      & $1.93$ & $0.28$ & $85.31$\% \\
BakLLAVA~\cite{liu2023improvedllava}    & $2.29$ & $0.40$ & $82.50$\% \\
BLIP~\cite{li2022blip}        & $0.67$ & $0.44$ & $34.60$\% \\
FC ~\cite{rennie2017self}         & $0.36$ & $0.23$ & $35.75$\% \\
Florence~\cite{florence}     & $0.40$ & $0.15$ & $63.58$\% \\
LLAVA~\cite{liu2023llava}        & $0.48$ & $0.24$ & $50.49$\% \\
Oscar~\cite{li2020oscar}       & $0.46$ & $0.40$ & $14.90$\% \\
SAT~\cite{xu2015show}         & $0.41$ & $0.19$ & $54.47$\% \\
NIC~\cite{vinyals2015show}         & $1.01$ & $0.32$ & $68.95$\% \\
NIC+Equal.~\cite{Hendricks_2018_ECCV}  & $7.46$ & $0.23$ & $96.91$\% \\
Transformer~\cite{Attention_is_all_you_need} & $6.81$ & $0.23$ & $96.61$\% \\
UpDown~\cite{UpDn}      & $0.50$ & $0.20$ & $60.50$\% \\
VIT\_GPT2~\cite{vit_gpt2}   & $0.54$ & $0.39$ & $29.18$\% \\
\midrule
\textbf{Average} & -- & -- & $59.52$\% \\
\bottomrule
\end{tabularx}
\caption{Sentence encoders trained from scratch}
\label{tab:scratch_attackers_race}
\end{subtable}

\begin{subtable}[t]{0.35\textwidth}
\centering
\begin{tabularx}{\textwidth}{cccc}
\toprule
\makecell{\textbf{Captioning} \\ \textbf{Models}} & \makecell{\textbf{CV}\\\textbf{(LIC)}} & \makecell{\textbf{CV}\\\textbf{(DBAC)}} & \makecell{\textbf{$\%$}\\ \textbf{Reduction}}\\
\midrule
Att2In      &  $0.97$ & $0.16$ &  $83.81$\% \\
BakLLAVA    &  $2.24$ & $0.09$ &  $96.15\%$ \\
BLIP        &  $1.37$ & $0.61$ & $55.27\%$ \\
FC          &  $3.16$ & $0.24$ &  $92.25\%$ \\
Florence    &  $1.06$ & $0.03$ &  $96.94\%$ \\
LLAVA       &  $18.20$ \ \ & $0.08$ &  $99.54\%$ \\
Oscar       &  $1.46$ & $0.58$ &  $60.62\%$ \\
SAT         &  $1.85$ & $0.43$ &  $76.80\%$ \\
NIC         &  $5.24$ & $0.42$ &  $92.00\%$ \\
NIC+Equal.  &  $22.06$ \ \ & $0.60$ &  $97.29\%$ \\
Transformer &  $2.05$ & $0.29$ &  $85.65\%$ \\
UpDown      &  $1.02$ & $0.31$ &  $70.06\%$ \\
VIT\_GPT2   &  $1.77$ & $0.35$ &  $80.34\%$ \\
\midrule
\textbf{Average} & -- & -- & $83.60\%$ \\
\bottomrule
\end{tabularx}
\caption{Pre-trained sentence encoders}
\label{tab:pretrained_attackers_race}
\end{subtable}
\vspace{-0.2cm}
\caption{Sensitivity of DBAC and LIC to different sentence encoders (race results): For the encoders trained from scratch, the CV in DBAC was $59.5\%$ less than LIC. For pre-trained encoders, the CV in DBAC was $83.6\%$ less than LIC.} 
\label{tab:attackers_comparison_race}
\vspace{-0.4cm}
\end{table}



\subsection{Sensitivity to Sentence Encoders}
\label{subsec:sentence_encoder_sensitivity}
\subsubsection{Experiment Setup}

We examined whether LIC and DBAC reported consistent bias amplification scores across sentence encoders (e.g., LSTM, RNN). With a consistent metric, users need not worry about choosing a suitable encoder to measure bias amplification. We did not test metrics like DPA and $BA_{\rightarrow}$ as they do not use sentence encoders.

For each captioning model, we calculated the coefficient of variation (CV) in DBAC and LIC scores across $6$ sentence encoders trained from scratch on COCO captions, and across $4$ pre-trained encoders. Since pre-trained encoders carry biases from their training data \cite{sentence_encoder_bias}, we computed CV across them separately. CV was chosen over variance as it is scale invariant. A low CV is desirable as it indicates a consistent metric.

The $6$ sentence encoders trained from scratch on COCO captions: LSTM-Unidirectional \cite{hochreiter1997long}, LSTM-Bidirectional \cite{hochreiter1997long}, RNN-Unidirectional \cite{schmidt2019recurrent}, RNN-Bidirectional \cite{schmidt2019recurrent}, TransformerEncoder (1 head) \cite{Attention_is_all_you_need}, and TransformerEncoder (5 heads) \cite{Attention_is_all_you_need}. The $4$ pre-trained encoders: MiniLM-L6-v2~\cite{MiniLM}, Mpnet-base-v2~\cite{song2020mpnet}, Distilroberta-v1~\cite{distilroberta}, and Distilbert~\cite{distilroberta}. Each sentence encoder was paired with a 3-layer MLP to form the attacker model. For DBAC, we measured $T \rightarrow A$ bias amplification.

\subsubsection{Results}
\textbf{For gender and race, DBAC was much more consistent than LIC across sentence encoders.} For gender, on each captioning model, we showed the CV values for LIC and DBAC across the six sentence encoders trained from scratch on COCO captions (Table \ref{tab:scratch_attackers}). On average, DBAC showed $84.5\%$ less variation in bias amplification scores compared to LIC. In Table \ref{tab:pretrained_attackers}, we showed the CV values across the four pre-trained encoders. DBAC showed $69.9\%$ less variation in scores compared to LIC. 

For race, we showed the CV values across the six sentence encoders trained from scratch in Table \ref{tab:scratch_attackers_race}. DBAC showed $59.5\%$ less variation in scores compared to LIC. In Table \ref{tab:pretrained_attackers_race}, we showed the CV values across the four pre-trained encoders. DBAC showed $83.6\%$ less variation in scores compared to LIC. 

Compared to LIC, DBAC is consistent across encoders. DBAC eliminates the pain of choosing a suitable encoder to measure bias amplification. We reported the complete set of bias amplification scores for DBAC and LIC, across all sentence encoders in Section~\ref{sec: Sensitivity Complete Table}.

\subsection{Constant vs. Contextual Substitution}
\label{subsec:constant_vs_contextual}
\subsubsection{Experiment Setup}
We compared the two vocabulary alignment techniques discussed in Section \ref{subsec:vocab_align}: constant and contextual substitution. We examined which substitution technique better retained the original meaning of the HGCs. 

For each model, we computed two similarity scores: (1) between the original HGC and the HGC after constant substitution, and (2) between the original HGC and the HGC after contextual substitution. We used METEOR \cite{denkowski-lavie-2014-meteor} (a sentence similarity metric) to compute similarity scores.

We also evaluated how the bias in the HGCs changed for LIC  (reported using $\text{LIC}_{H}$) and for DBAC (reported using $\omega_{H}$), when we used contextual instead of constant substitution. For DBAC, we measured $A \rightarrow T$ bias. For contextual substitution, we tested two word embedding models: GloVe \cite{pennington-etal-2014-glove} and FastText \cite{bojanowski2016enriching}. We did not study $BA_{\rightarrow}$ and DPA as they do not use vocabulary alignment. We showed gender results here, and race results in Section \ref{sec: Race Experiments}.

\begin{table}[htbp]     
    \centering     
    \footnotesize     
    \begin{tabular}{P{1cm}P{1cm}P{1.8cm}P{1.8cm}}     
    \toprule     
    \textbf{Models} & \textbf{Constant} & \multicolumn{2}{c}{\textbf{Contextual}} \\     
    \cmidrule(lr){3-4}      
     & & \textbf{GloVe} & \textbf{FastText} \\     
    \midrule     
    Att2In      & $0.631$ & $0.655$ (\green{$3.82$\%$\uparrow$}) & $0.663$ (\green{$4.94$\%$\uparrow$})\\     
    BakLLAVA    & $0.825$ & $0.829$ (\green{$0.45$\%$\uparrow$}) & $0.830$ (\green{$0.61$\%$\uparrow$})\\     
    BLIP        & $0.779$ & $0.794$ (\green{$1.98$\%$\uparrow$}) & $0.798$ (\green{$2.45$\%$\uparrow$})\\     
    FC          & $0.574$ & $0.598$ (\green{$4.09$\%$\uparrow$}) & $0.604$ (\green{$5.19$\%$\uparrow$})\\     
    Florence    & $0.805$ & $0.819$ (\green{$1.54$\%$\uparrow$}) & $0.821$ (\green{$1.99$\%$\uparrow$})\\     
    LLAVA       & $0.816$ & $0.822$ (\green{$0.77$\%$\uparrow$}) & $0.824$ (\green{$1.08$\%$\uparrow$})\\     
    Oscar       & $0.629$ & $0.650$ (\green{$3.26$\%$\uparrow$}) & $0.657$ (\green{$4.34$\%$\uparrow$})\\     
    SAT         & $0.713$ & $0.738$ (\green{$3.54$\%$\uparrow$}) & $0.741$ (\green{$3.94$\%$\uparrow$})\\     
    NIC         & $0.693$ & $0.723$ (\green{$4.24$\%$\uparrow$}) & $0.727$ (\green{$4.83$\%$\uparrow$})\\     
    NIC+Equal.  & $0.731$ & $0.752$ (\green{$2.85$\%$\uparrow$}) & $0.757$ (\green{$3.54$\%$\uparrow$})\\     
    Transformer & $0.762$ & $0.776$ (\green{$1.84$\%$\uparrow$}) & $0.779$ (\green{$2.30$\%$\uparrow$})\\     
    UpDown      & $0.658$ & $0.688$ (\green{$4.45$\%$\uparrow$}) & $0.693$ (\green{$5.32$\%$\uparrow$})\\     
    VIT\_GPT2   & $0.741$ & $0.759$ (\green{$2.42$\%$\uparrow$}) & $0.765$ (\green{$3.23$\%$\uparrow$})\\     
    \midrule     
    \textbf{Average} & $0.720$ & $0.738$ (\green{$2.71$\%$\uparrow$}) & $0.743$ (\green{$3.37$\%$\uparrow$})\\     
    \bottomrule     
    \end{tabular}     
    \vspace{-0.2cm}     
    \caption{METEOR scores for constant vs. contextual substitution: Values in green indicate $\%$ increase in METEOR scores with contextual substitution, relative to constant substitution. For all captioning models, contextual substitution achieved a higher METEOR score than constant substitution.}     
    \vspace{-0.5cm}     
    \label{tab: Meteor Constant vs. Context} 
\end{table}

\begin{table*}[htbp]
\centering
\footnotesize
\begin{subtable}[t]{0.43\textwidth}
\centering
\begin{tabularx}{\textwidth}{P{1cm}P{0.9cm}P{2cm}P{2cm}}
\toprule
\textbf{Models} & \textbf{Constant} & \multicolumn{2}{c}{\textbf{Contextual}} \\
\cmidrule(lr){3-4}
 & & \textbf{GloVe} & \textbf{FastText} \\
\midrule
Att2In      & $0.382$ & $0.402$ (\green{$5.39$\% $\uparrow$}) & $0.400$ (\green{$4.76$\% $\uparrow$})\\
BakLLAVA    & $0.389$ & $0.403$ (\green{$3.78$\% $\uparrow$}) & $0.425$ (\green{$9.37$\% $\uparrow$})\\
BLIP        & $0.380$ & $0.400$ (\green{$5.10$\% $\uparrow$}) & $0.401$ (\green{$5.42$\% $\uparrow$})\\
FC          & $0.381$ & $0.400$ (\green{$5.06$\% $\uparrow$}) & $0.401$ (\green{$5.27$\% $\uparrow$})\\
Florence    & $0.397$ & $0.401$ (\green{$0.96$\% $\uparrow$}) & $0.400$ (\green{$0.76$\% $\uparrow$})\\
LLAVA       & $0.378$ & $0.400$ (\green{$5.74$\% $\uparrow$}) & $0.404$ (\green{$6.85$\% $\uparrow$})\\
Oscar       & $0.421$ & $0.432$ (\green{$2.74$\% $\uparrow$}) & \ $0.477$ (\green{$13.44$\%$\uparrow$})\\
SAT         & $0.391$ & $0.400$ (\green{$2.22$\% $\uparrow$}) & $0.401$ (\green{$2.56$\% $\uparrow$})\\
NIC         & $0.377$ & $0.400$ (\green{$6.26$\% $\uparrow$}) & $0.401$ (\green{$6.42$\% $\uparrow$})\\
NIC+Equal.  & $0.399$ & $0.401$ (\green{$0.40$\% $\uparrow$}) & $0.404$ (\green{$1.28$\% $\uparrow$})\\
Transformer & $0.453$ & $0.457$ (\green{$0.88$\% $\uparrow$}) & $0.461$ (\green{$1.68$\% $\uparrow$})\\
UpDown      & $0.399$ & $0.400$ (\green{$0.25$\% $\uparrow$}) & \ $0.456$ (\green{$14.32$\%$\uparrow$})\\
VIT\_GPT2   & $0.463$ & $0.483$ (\green{$4.54$\% $\uparrow$}) & $0.470$ (\green{$1.60$\% $\uparrow$})\\
\midrule
\textbf{Average} & $0.401$ & $0.414$ (\green{$3.33$\% $\uparrow$}) & $0.423$ (\green{$5.67$\% $\uparrow$})\\
\bottomrule
\end{tabularx}
\caption{DBAC results for the HGCs: $\omega_{H:A \rightarrow T}$}
\label{tab:DBAC Constant vs. Context}
\end{subtable}
\hfill
\begin{subtable}[t]{0.44\textwidth}
\centering
\begin{tabularx}{\textwidth}{P{1cm}P{0.9cm}P{2cm}P{2.1cm}}
\toprule
\textbf{Models} & \textbf{Constant} & \multicolumn{2}{c}{\textbf{Contextual}} \\
\cmidrule(lr){3-4}
 & & \textbf{GloVe} & \textbf{FastText} \\
\midrule
Att2In      & $0.595$ & $0.619$ (\green{$4.12$\%$\uparrow$}) & $0.630$ (\green{$5.98$\%$\uparrow$}) \\
BakLLAVA    & $0.851$ & $0.877$ (\green{$3.00$\% $\uparrow$}) & $0.898$ (\green{$5.42$\%$\uparrow$}) \\
BLIP        & $0.732$ & $0.751$ (\green{$2.63$\% $\uparrow$}) & $0.759$ (\green{$3.76$\%$\uparrow$}) \\
FC          & $0.575$ & $0.625$ (\green{$8.75$\% $\uparrow$}) & \ \ $0.633$ (\green{$10.01$\%$\uparrow$}) \\
Florence    & $0.822$ & $0.848$ (\green{$3.22$\% $\uparrow$}) & $0.867$ (\green{$5.32$\%$\uparrow$}) \\
LLAVA       & $0.816$ & $0.837$ (\green{$2.64$\% $\uparrow$}) & $0.846$ (\green{$3.72$\%$\uparrow$})\\
Oscar       & $0.649$ & $0.688$ (\green{$5.96$\% $\uparrow$}) & $0.705$ (\green{$8.69$\%$\uparrow$})\\
SAT         & $0.611$ & $0.636$ (\green{$4.09$\% $\uparrow$}) & $0.660$ (\green{$8.07$\%$\uparrow$}) \\
NIC         & $0.699$ & $0.742$ (\green{$6.12$\% $\uparrow$}) & $0.755$ (\green{$8.05$\%$\uparrow$}) \\
NIC+Equal.  & $0.591$ & $0.598$ (\green{$1.25$\% $\uparrow$}) & $0.610$ (\green{$3.25$\%$\uparrow$}) \\
Transformer & $0.597$ & $0.601$ (\green{$0.70$\% $\uparrow$}) & $0.615$ (\green{$2.95$\%$\uparrow$}) \\
UpDown      & $0.621$ & $0.648$ (\green{$4.26$\% $\uparrow$}) & $0.661$ (\green{$6.31$\%$\uparrow$}) \\
VIT\_GPT2   & $0.543$ & $0.594$ (\green{$9.37$\% $\uparrow$}) & \ \ $0.609$ (\green{$12.10$\%$\uparrow$}) \\
\midrule
\textbf{Average} & $0.669$ & $0.697$ (\green{$4.31$\% $\uparrow$}) & $0.711$ (\green{$6.43$\%$\uparrow$})\\
\bottomrule
\end{tabularx}
\caption{LIC results for the HGCs: $LIC_{H}$}
\label{tab:LIC Constant vs. Context}
\end{subtable}
\vspace{-0.3cm}
\caption{DBAC and LIC scores for constant vs. contextual substitution: Values in green indicate the $\%$ increase in HGC's bias reported by contextual substitution relative to constant substitution. Across all captioning models, both DBAC and LIC consistently reported higher bias in the HGCs with contextual substitution.}
\vspace{-0.3cm}
\label{tab: Constant vs. Contextual Substituition}
\end{table*}

\subsubsection{Results}
\label{subsec:vocab_substi_results}
In column $2$ of Table \ref{tab: Meteor Constant vs. Context}, we showed the METEOR similarity scores for constant substitution. Columns $3$ and $4$ show the METEOR scores for contextual substitution with GloVe and FastText, respectively. The percentage values in green indicate the $\%$ increase in METEOR score achieved by contextual substitution, relative to constant substitution.

\textbf{Contextual substitution better retained the original meaning of the HGCs.} Across all captioning models, contextual substitution produced higher similarity scores than constant substitution. On average, contextual substitution improved the METEOR sentence similarity score by $2.7\%$ with GloVe, and $3.4\%$ with FastText. 

In column $2$ of Table \ref{tab:DBAC Constant vs. Context}, we showed DBAC scores for the HGCs ($\omega_{H: A \rightarrow T}$) with constant substitution. Columns $3$ and $4$ show the DBAC scores for contextual substitution with GloVe and FastText, respectively. The percentage values in green indicate the $\%$ increase in DBAC score achieved by contextual substitution, relative to constant substitution. We created an identical table for LIC scores in Table \ref{tab:LIC Constant vs. Context}.

\textbf{DBAC and LIC reported higher bias in the HGCs with contextual substitution.} On average, contextual substitution increased DBAC scores by $3.3\%$ with GloVe, and $5.7\%$ with FastText (Table \ref{tab:DBAC Constant vs. Context}). Contextual substitution increased LIC scores by $4.3\%$ with GloVe, and $6.4\%$ with FastText (Table \ref{tab:LIC Constant vs. Context}).

We observed similar insights for race (see Section \ref{sec: Race Experiments}). These results are intuitive. Consider two HGCs: ``a $<$gender$>$ changing a tire.'' and ``a $<$gender$>$ changing a diaper.'' Assume the words tire and diaper are not present in $V_{MGC}$. In constant substitution, these captions become the same: a $<$gender$>$ changing a $<$unk$>$. We lost meaningful words, or the bias signal, that could distinguish these captions (tire vs. diaper). Contextual substitution retains this bias signal (e.g., replacing tire with wheel, and diaper with underpants). This explains why contextual substitution reported a higher bias in HGCs, with both LIC and DBAC.

\begin{table}[htbp]
    \centering
    \footnotesize
    \vspace{-0.2cm}
    \begin{tabular}{P{1.5cm}cc}
        \toprule
        \textbf{Model} &\multicolumn{2}{c}{\textbf{\ABB}}\\
        \cmidrule(lr){2-3}
         & $\textbf{A} \rightarrow \textbf{T}$ & $\textbf{T} \rightarrow \textbf{A}$\\
         \midrule
         Att2In & $-40.05_{0.13}$ & $-05.89_{0.33}$\\
         BakLLAVA & $-62.97_{0.06}$ & $-02.98_{0.18}$\\
         BLIP & $-33.85_{0.07}$ &  \ \ \ $09.98_{0.33}$\\
         FC & $-39.39_{0.14}$ & \ \ \ $06.45_{0.24}$\\
         Florence & $-57.89_{0.07}$ & $-10.25_{0.37}$\\
         LLAVA & $-76.06_{0.06}$ & $-06.76_{0.23}$\\
         Oscar & $-40.02_{0.16}$ & \ \ \ $11.19_{0.23}$\\
         SAT & $-30.11_{0.13}$ & \ \ \ $07.61_{0.12}$\\
         NIC & $-43.66_{0.19}$ & \ \ \ $27.43_{0.56}$\\
         NIC+Equal. & $-08.87_{0.12}$ & \ \ \ $25.52_{0.43}$\\
         Transformer & $-36.55_{0.21}$ & \ \ \ $12.15_{0.16}$\\ 
         UpDown & $-41.79_{0.17}$ & \ \ \ $11.72_{0.25}$\\
         ViT\_GPT2 & $-38.54_{0.40}$ & $-02.06_{0.36}$\\         
         \bottomrule
    \end{tabular}
    \vspace{-0.2cm}
    \caption{For $13$ captioning models, we measured $A \rightarrow T$ and $T \rightarrow A$ bias amplification using DBAC. Values are scaled by $100$. Subscripts show 95\% confidence intervals.}
    \label{tab:overall}
    \vspace{-0.6cm}
\end{table}

\section{Discussion}
Our results show that DBAC is the only reliable metric to measure bias amplification in captioning models. LIC gives partial information about bias amplification as it is not directional. General directional metrics like $BA_{\rightarrow}$ and DPA are unreliable, as they cannot capture language biases in captions. In this section, we used DBAC to discuss insights on bias amplification in $13$ captioning models, trained on COCO's HGCs (note that BakLLAVA, LLAVA, and Florence were not trained on COCO).

\textbf{Mitigation techniques like Equalizer worsened $\mathbf{A \rightarrow T}$ bias in COCO's HGCs.} In Table \ref{tab:overall}, the $T \rightarrow A$ bias amplification in NIC+Equalizer is lower than NIC. However, the $A \rightarrow T$ bias amplification is higher. Equalizer (a bias mitigation technique) mitigated biases in $T \rightarrow A$. However, it unintentionally made the bias worse in $A \rightarrow T$. 

\textbf{We need to make captioning models fairer at predicting gender, given task objects.} All $13$ models reduced $A \rightarrow T$ biases, but $8$ of them increased $T \rightarrow A$ biases. A reduction in $A \rightarrow T$ shows that, given a gender, all models became fairer in predicting tasks. An increase in $T \rightarrow A$ shows that given a task, $8$ models became more biased in predicting gender.

\vspace{-0.2cm}
\section{Limitations}

DBAC (and LIC) use sentence encoders to quantify biases. In some cases, sentence encoders can wrongly quantify biases due to instabilities (e.g., the encoder being trapped in local minima during training). To overcome these instabilities, bias amplification values are computed over multiple iterations (using confidence intervals) to ensure reliable results. DBAC (and LIC) are computationally expensive than co-occurrence-based metrics like $BA_{\rightarrow}$.

In DBAC, we performed contextual substitution using pre-trained word embedding models like GloVe \cite{pennington-etal-2014-glove}, and FastText \cite{bojanowski2016enriching}. Prior studies have shown that these word embedding models carry biases from their training dataset \cite{bolukbasi2016man,VeraExploringAM}. These biases can influence DBAC scores. We recommend running DBAC with multiple word embedding models to ensure unbiased scores. 

\vspace{-0.2cm}
\section{Conclusion}

We introduced DBAC, a language-aware and directional metric to measure bias amplification in captioning models. DBAC can reliably identify when captioning models amplify biases. DBAC offers several other improvements over LIC (the previous state-of-the-art metric): (1) it is minimally sensitive to sentence encoders. With DBAC, users can use any encoder to measure bias amplification in their captioning model (this was not possible with LIC). (2) DBAC offers a more accurate estimate of bias amplification as it uses contextual substitution for vocabulary alignment. 
{
    \small
    \bibliographystyle{ieeenat_fullname}
    \bibliography{main}

@String(CVPR= {IEEE Conf. Comput. Vis. Pattern Recog.})

@String(ICCV= {Int. Conf. Comput. Vis.})

@String(ECCV= {Eur. Conf. Comput. Vis.})

@String(NIPS= {Adv. Neural Inform. Process. Syst.})

@String(AAAI = {AAAI})

@String(CVPR  = {CVPR})

@String(ICCV  = {ICCV})

@String(ECCV  = {ECCV})

@String(NIPS  = {NeurIPS})

@INPROCEEDINGS {LIC,
author = {Y. Hirota and Y. Nakashima and N. Garcia},
booktitle = {2022 IEEE/CVF Conference on Computer Vision and Pattern Recognition (CVPR)},
title = {Quantifying Societal Bias Amplification in Image Captioning},
year = {2022},
volume = {},
issn = {},
pages = {13440-13449},
doi = {10.1109/CVPR52688.2022.01309},
url = {https://doi.ieeecomputersociety.org/10.1109/CVPR52688.2022.01309},
publisher = {IEEE Computer Society},
address = {Los Alamitos, CA, USA},
month = {jun}
}

@inproceedings{DBA,
title = "Directional Bias Amplification",
author = "Angelina Wang and Olga Russakovsky",
note = "Publisher Copyright: Copyright {\textcopyright} 2021 by the author(s); 38th International Conference on Machine Learning, ICML 2021 ; Conference date: 18-07-2021 Through 24-07-2021",
year = "2021",
language = "English (US)",
series = "Proceedings of Machine Learning Research",
publisher = "ML Research Press",
pages = "10882--10893",
booktitle = "Proceedings of the 38th International Conference on Machine Learning, ICML 2021",
}

@article{Image_Cap_2, title={ImageCaptioner2: Image Captioner for Image Captioning Bias Amplification Assessment}, volume={38}, url={https://ojs.aaai.org/index.php/AAAI/article/view/30080}, DOI={10.1609/aaai.v38i19.30080}, abstractNote={Most pre-trained learning systems are known to suffer from bias, which typically emerges from the data, the model, or both. Measuring and quantifying bias and its sources is a challenging task and has been extensively studied in image captioning. Despite the significant effort in this direction, we observed that existing metrics lack consistency in the inclusion of the visual signal. In this paper, we introduce a new bias assessment metric, dubbed ImageCaptioner2, for image captioning. Instead of measuring the absolute bias in the model or the data, ImageCaptioner2pay more attention to the bias introduced by the model w.r.t the data bias, termed bias amplification. Unlike the existing methods, which only evaluate the image captioning algorithms based on the generated captions only, ImageCaptioner2incorporates the image while measuring the bias. In addition, we design a formulation for measuring the bias of generated captions as prompt-based image captioning instead of using language classifiers. Finally, we apply our ImageCaptioner2metric across 11 different image captioning architectures on three different datasets, i.e., MS-COCO caption dataset, Artemis V1, and Artemis V2, and on three different protected attributes, i.e., gender, race, and emotions. Consequently, we verify the effectiveness of our ImageCaptioner2metric by proposing Anonymous-Bench, which is a novel human evaluation paradigm for bias metrics. Our metric shows significant superiority over the recent bias metric; LIC, in terms of human alignment, where the correlation scores are 80% and 54% for our metric and LIC, respectively. The code and more details are available at https://eslambakr.github.io/imagecaptioner2.github.io/.}, number={19}, journal={Proceedings of the AAAI Conference on Artificial Intelligence}, author={Abdelrahman, Eslam and Sun, Pengzhan and Li, Li Erran and Elhoseiny, Mohamed}, year={2024}, month={Mar.}, pages={20902-20911} }

@inproceedings{zhao-etal-2017-men,
    title = "Men Also Like Shopping: Reducing Gender Bias Amplification using Corpus-level Constraints",
    author = "Zhao, Jieyu  and
      Wang, Tianlu  and
      Yatskar, Mark  and
      Ordonez, Vicente  and
      Chang, Kai-Wei",
    editor = "Palmer, Martha  and
      Hwa, Rebecca  and
      Riedel, Sebastian",
    booktitle = "Proceedings of the 2017 Conference on Empirical Methods in Natural Language Processing",
    month = sep,
    year = "2017",
    address = "Copenhagen, Denmark",
    publisher = "Association for Computational Linguistics",
    url = "https://aclanthology.org/D17-1323",
    doi = "10.18653/v1/D17-1323",
    pages = "2979--2989",
}

@InProceedings{MDBA,
  title = 	 {Men Also Do Laundry: Multi-Attribute Bias Amplification},
  author =       {Zhao, Dora and Andrews, Jerone and Xiang, Alice},
  booktitle = 	 {Proceedings of the 40th International Conference on Machine Learning},
  pages = 	 {42000--42017},
  year = 	 {2023},
  editor = 	 {Krause, Andreas and Brunskill, Emma and Cho, Kyunghyun and Engelhardt, Barbara and Sabato, Sivan and Scarlett, Jonathan},
  volume = 	 {202},
  series = 	 {Proceedings of Machine Learning Research},
  month = 	 {23--29 Jul},
  publisher =    {PMLR},
  url = 	 {https://proceedings.mlr.press/v202/zhao23a.html}
}

@inproceedings{Attention_is_all_you_need,
 author = {Vaswani, Ashish and Shazeer, Noam and Parmar, Niki and Uszkoreit, Jakob and Jones, Llion and Gomez, Aidan N and Kaiser, \L ukasz and Polosukhin, Illia},
 booktitle = {Advances in Neural Information Processing Systems},
 editor = {I. Guyon and U. Von Luxburg and S. Bengio and H. Wallach and R. Fergus and S. Vishwanathan and R. Garnett},
 pages = {},
 publisher = {Curran Associates, Inc.},
 title = {Attention is All you Need},
 url = {https://proceedings.neurips.cc/paper_files/paper/2017/file/3f5ee243547dee91fbd053c1c4a845aa-Paper.pdf},
 volume = {30},
 year = {2017}
}

@INPROCEEDINGS {UpDn,
author = {P. Anderson and X. He and C. Buehler and D. Teney and M. Johnson and S. Gould and L. Zhang},
booktitle = {2018 IEEE/CVF Conference on Computer Vision and Pattern Recognition (CVPR)},
title = {Bottom-Up and Top-Down Attention for Image Captioning and Visual Question Answering},
year = {2018},
volume = {},
issn = {},
pages = {6077-6086},
doi = {10.1109/CVPR.2018.00636},
url = {https://doi.ieeecomputersociety.org/10.1109/CVPR.2018.00636},
publisher = {IEEE Computer Society},
address = {Los Alamitos, CA, USA},
month = {jun}
}

@misc{lin2019crankvolumepreferencebias,
      title={Crank up the volume: preference bias amplification in collaborative recommendation}, 
      author={Kun Lin and Nasim Sonboli and Bamshad Mobasher and Robin Burke},
      year={2019},
      eprint={1909.06362},
      archivePrefix={arXiv},
      primaryClass={cs.IR},
      url={https://arxiv.org/abs/1909.06362}, 
}

@inproceedings{seshadri-etal-2024-bias,
    title = "The Bias Amplification Paradox in Text-to-Image Generation",
    author = "Seshadri, Preethi  and
      Singh, Sameer  and
      Elazar, Yanai",
    editor = "Duh, Kevin  and
      Gomez, Helena  and
      Bethard, Steven",
    booktitle = "Proceedings of the 2024 Conference of the North American Chapter of the Association for Computational Linguistics: Human Language Technologies (Volume 1: Long Papers)",
    month = jun,
    year = "2024",
    address = "Mexico City, Mexico",
    publisher = "Association for Computational Linguistics",
    url = "https://aclanthology.org/2024.naacl-long.353",
    doi = "10.18653/v1/2024.naacl-long.353",
    pages = "6367--6384",
}

@INPROCEEDINGS{intersectional_defintion_of_fairness,
  author={Foulds, James R. and Islam, Rashidul and Keya, Kamrun Naher and Pan, Shimei},
  booktitle={2020 IEEE 36th International Conference on Data Engineering (ICDE)}, 
  title={An Intersectional Definition of Fairness}, 
  year={2020},
  volume={},
  number={},
  pages={1918-1921},
  doi={10.1109/ICDE48307.2020.00203}
}

@InProceedings{Hendricks_2018_ECCV,
author = {Hendricks, Lisa Anne and Burns, Kaylee and Saenko, Kate and Darrell, Trevor and Rohrbach, Anna},
title = {Women also Snowboard: Overcoming Bias in Captioning Models},
booktitle = {Proceedings of the European Conference on Computer Vision (ECCV)},
month = {September},
year = {2018}
}

@inproceedings{VeraExploringAM,
  title={Exploring and Mitigating Gender Bias in GloVe Word Embeddings},
  author={Mauro Vera},
  url={https://api.semanticscholar.org/CorpusID:4806521},
  year={2018}
}

@article{bolukbasi2016man,
  title={Man is to computer programmer as woman is to homemaker? debiasing word embeddings},
  author={Bolukbasi, Tolga and Chang, Kai-Wei and Zou, James Y and Saligrama, Venkatesh and Kalai, Adam T},
  journal={Advances in neural information processing systems},
  volume={29},
  year={2016}
}

@inproceedings{pennington-etal-2014-glove,
    title = "{G}lo{V}e: Global Vectors for Word Representation",
    author = "Pennington, Jeffrey  and
      Socher, Richard  and
      Manning, Christopher",
    editor = "Moschitti, Alessandro  and
      Pang, Bo  and
      Daelemans, Walter",
    booktitle = "Proceedings of the 2014 Conference on Empirical Methods in Natural Language Processing ({EMNLP})",
    month = oct,
    year = "2014",
    address = "Doha, Qatar",
    publisher = "Association for Computational Linguistics",
    url = "https://aclanthology.org/D14-1162/",
    doi = "10.3115/v1/D14-1162",
    pages = "1532--1543"
}

@inproceedings{meister2023gender,
  title={Gender artifacts in visual datasets},
  author={Meister, Nicole and Zhao, Dora and Wang, Angelina and Ramaswamy, Vikram V and Fong, Ruth and Russakovsky, Olga},
  booktitle={Proceedings of the IEEE/CVF International Conference on Computer Vision},
  pages={4837--4848},
  year={2023}
}

@misc{tokas2024makingbiasamplificationbalanced,
      title={Making Bias Amplification in Balanced Datasets Directional and Interpretable}, 
      author={Bhanu Tokas and Rahul Nair and Hannah Kerner},
      year={2024},
      eprint={2412.11060},
      archivePrefix={arXiv},
      primaryClass={cs.CV},
      url={https://arxiv.org/abs/2412.11060}, 
}

@article{chen2015microsoft,
  title={Microsoft coco captions: Data collection and evaluation server},
  author={Chen, Xinlei and Fang, Hao and Lin, Tsung-Yi and Vedantam, Ramakrishna and Gupta, Saurabh and Doll{\'a}r, Piotr and Zitnick, C Lawrence},
  journal={arXiv preprint arXiv:1504.00325},
  year={2015}
}

@inproceedings{rennie2017self,
  title={Self-critical sequence training for image captioning},
  author={Rennie, Steven J and Marcheret, Etienne and Mroueh, Youssef and Ross, Jerret and Goel, Vaibhava},
  booktitle={Proceedings of the IEEE conference on computer vision and pattern recognition},
  pages={7008--7024},
  year={2017}
}

@inproceedings{li2020oscar,
  title={Oscar: Object-semantics aligned pre-training for vision-language tasks},
  author={Li, Xiujun and Yin, Xi and Li, Chunyuan and Zhang, Pengchuan and Hu, Xiaowei and Zhang, Lei and Wang, Lijuan and Hu, Houdong and Dong, Li and Wei, Furu and others},
  booktitle={Computer Vision--ECCV 2020: 16th European Conference, Glasgow, UK, August 23--28, 2020, Proceedings, Part XXX 16},
  pages={121--137},
  year={2020},
  organization={Springer}
}

@inproceedings{xu2015show,
  title={Show, attend and tell: Neural image caption generation with visual attention},
  author={Xu, Kelvin and Ba, Jimmy and Kiros, Ryan and Cho, Kyunghyun and Courville, Aaron and Salakhudinov, Ruslan and Zemel, Rich and Bengio, Yoshua},
  booktitle={International conference on machine learning},
  pages={2048--2057},
  year={2015},
  organization={PMLR}
}

@inproceedings{vinyals2015show,
  title={Show and tell: A neural image caption generator},
  author={Vinyals, Oriol and Toshev, Alexander and Bengio, Samy and Erhan, Dumitru},
  booktitle={Proceedings of the IEEE conference on computer vision and pattern recognition},
  pages={3156--3164},
  year={2015}
}

@inproceedings{denkowski-lavie-2014-meteor,
    title = "Meteor Universal: Language Specific Translation Evaluation for Any Target Language",
    author = "Denkowski, Michael  and
      Lavie, Alon",
    editor = "Bojar, Ond{\v{r}}ej  and
      Buck, Christian  and
      Federmann, Christian  and
      Haddow, Barry  and
      Koehn, Philipp  and
      Monz, Christof  and
      Post, Matt  and
      Specia, Lucia",
    booktitle = "Proceedings of the Ninth Workshop on Statistical Machine Translation",
    month = jun,
    year = "2014",
    address = "Baltimore, Maryland, USA",
    publisher = "Association for Computational Linguistics",
    url = "https://aclanthology.org/W14-3348/",
    doi = "10.3115/v1/W14-3348",
    pages = "376--380"
}

@inproceedings{li2022blip,
      title={BLIP: Bootstrapping Language-Image Pre-training for Unified Vision-Language Understanding and Generation}, 
      author={Junnan Li and Dongxu Li and Caiming Xiong and Steven Hoi},
      year={2022},
      booktitle={ICML},
}

@INPROCEEDINGS{vit_gpt2,
  author={Mishra, Swapneel and Seth, Saumya and Jain, Shrishti and Pant, Vasudev and Parikh, Jolly and Jain, Rachna and Islam, Sardar M.N.},
  booktitle={2024 2nd International Conference on Advancement in Computation \& Computer Technologies (InCACCT)}, 
  title={Image Caption Generation using Vision Transformer and GPT Architecture}, 
  year={2024},
  volume={},
  number={},
  pages={1-6},
  doi={10.1109/InCACCT61598.2024.10551257}
}

@misc{liu2023llava,
      title={Visual Instruction Tuning}, 
      author={Liu, Haotian and Li, Chunyuan and Wu, Qingyang and Lee, Yong Jae},
      publisher={NeurIPS},
      year={2023},
}

@misc{liu2023improvedllava,
      title={Improved Baselines with Visual Instruction Tuning}, 
      author={Liu, Haotian and Li, Chunyuan and Li, Yuheng and Lee, Yong Jae},
      publisher={arXiv:2310.03744},
      year={2023},
}

@article{florence, title={Florence-2: Advancing a unified representation for a variety of vision tasks}, author={Xiao, Bin and Wu, Haiping and Xu, Weijian and Dai, Xiyang and Hu, Houdong and Lu, Yumao and Zeng, Michael and Liu, Ce and Yuan, Lu}, journal={arXiv preprint arXiv:2311.06242}, year={2023} 
}

@article{song2020mpnet,
    title={MPNet: Masked and Permuted Pre-training for Language Understanding},
    author={Song, Kaitao and Tan, Xu and Qin, Tao and Lu, Jianfeng and Liu, Tie-Yan},
    journal={arXiv preprint arXiv:2004.09297},
    year={2020}
}

@inproceedings{miniLM,
author = {Wang, Wenhui and Wei, Furu and Dong, Li and Bao, Hangbo and Yang, Nan and Zhou, Ming},
title = {MINILM: deep self-attention distillation for task-agnostic compression of pre-trained transformers},
year = {2020},
isbn = {9781713829546},
publisher = {Curran Associates Inc.},
address = {Red Hook, NY, USA},
booktitle = {Proceedings of the 34th International Conference on Neural Information Processing Systems},
articleno = {485},
numpages = {13},
location = {Vancouver, BC, Canada},
series = {NIPS '20}
}

@article{distilroberta,
  title={DistilBERT, a distilled version of BERT: smaller, faster, cheaper and lighter},
  author={Victor Sanh and Lysandre Debut and Julien Chaumond and Thomas Wolf},
  journal={ArXiv},
  year={2019},
  volume={abs/1910.01108}
}

@article{bojanowski2016enriching,
  title={Enriching Word Vectors with Subword Information},
  author={Bojanowski, Piotr and Grave, Edouard and Joulin, Armand and Mikolov, Tomas},
  journal={arXiv preprint arXiv:1607.04606},
  year={2016}
}

@INPROCEEDINGS{GradCam,
  author={Selvaraju, Ramprasaath R. and Cogswell, Michael and Das, Abhishek and Vedantam, Ramakrishna and Parikh, Devi and Batra, Dhruv},
  booktitle={2017 IEEE International Conference on Computer Vision (ICCV)}, 
  title={Grad-CAM: Visual Explanations from Deep Networks via Gradient-Based Localization}, 
  year={2017},
  volume={},
  number={},
  pages={618-626},
  doi={10.1109/ICCV.2017.74}}

@article{hochreiter1997long,
  title={Long short-term memory},
  author={Hochreiter, Sepp and Schmidhuber, J{\"u}rgen},
  journal={Neural computation},
  volume={9},
  number={8},
  pages={1735--1780},
  year={1997},
  publisher={MIT press}
}

@article{schmidt2019recurrent,
  title={Recurrent neural networks (rnns): A gentle introduction and overview},
  author={Schmidt, Robin M},
  journal={arXiv preprint arXiv:1912.05911},
  year={2019}
}

@article{tiwary2023accurate,
  title={An accurate generation of image captions for blind people using extended convolutional atom neural network},
  author={Tiwary, Tejal and Mahapatra, Rajendra Prasad},
  journal={Multimedia Tools and Applications},
  volume={82},
  number={3},
  pages={3801--3830},
  year={2023},
  publisher={Springer}
}

@article{zhang2023recognize,
  title={Recognize Anything: A Strong Image Tagging Model},
  author={Zhang, Youcai and Huang, Xinyu and Ma, Jinyu and Li, Zhaoyang and Luo, Zhaochuan and Xie, Yanchun and Qin, Yuzhuo and Luo, Tong and Li, Yaqian and Liu, Shilong and others},
  journal={arXiv preprint arXiv:2306.03514},
  year={2023}
}

@article{qiu2023gender,
  title={Gender biases in automatic evaluation metrics for image captioning},
  author={Qiu, Haoyi and Dou, Zi-Yi and Wang, Tianlu and Celikyilmaz, Asli and Peng, Nanyun},
  journal={arXiv preprint arXiv:2305.14711},
  year={2023}
}

@inproceedings{achlioptas2021artemis,
  title={Artemis: Affective language for visual art},
  author={Achlioptas, Panos and Ovsjanikov, Maks and Haydarov, Kilichbek and Elhoseiny, Mohamed and Guibas, Leonidas J},
  booktitle={Proceedings of the IEEE/CVF Conference on Computer Vision and Pattern Recognition},
  pages={11569--11579},
  year={2021}
}

@inproceedings{mohamed2022okay,
  title={It is okay to not be okay: Overcoming emotional bias in affective image captioning by contrastive data collection},
  author={Mohamed, Youssef and Khan, Faizan Farooq and Haydarov, Kilichbek and Elhoseiny, Mohamed},
  booktitle={Proceedings of the IEEE/CVF conference on computer vision and pattern recognition},
  pages={21263--21272},
  year={2022}
}

@inproceedings{sentence_encoder_bias,
    title = "On Measuring Social Biases in Sentence Encoders",
    author = "May, Chandler  and
      Wang, Alex  and
      Bordia, Shikha  and
      Bowman, Samuel R.  and
      Rudinger, Rachel",
    editor = "Burstein, Jill  and
      Doran, Christy  and
      Solorio, Thamar",
    booktitle = "Proceedings of the 2019 Conference of the North {A}merican Chapter of the Association for Computational Linguistics: Human Language Technologies, Volume 1 (Long and Short Papers)",
    month = jun,
    year = "2019",
    address = "Minneapolis, Minnesota",
    publisher = "Association for Computational Linguistics",
    url = "https://aclanthology.org/N19-1063/",
    doi = "10.18653/v1/N19-1063",
    pages = "622--628"
}
}
\clearpage
\appendix

\section{Sensitivity to Delta: GloVe Experiment}
\label{sec: glove thresholds}

\begin{figure}[h]
    \centering
    \includegraphics[width=\linewidth]{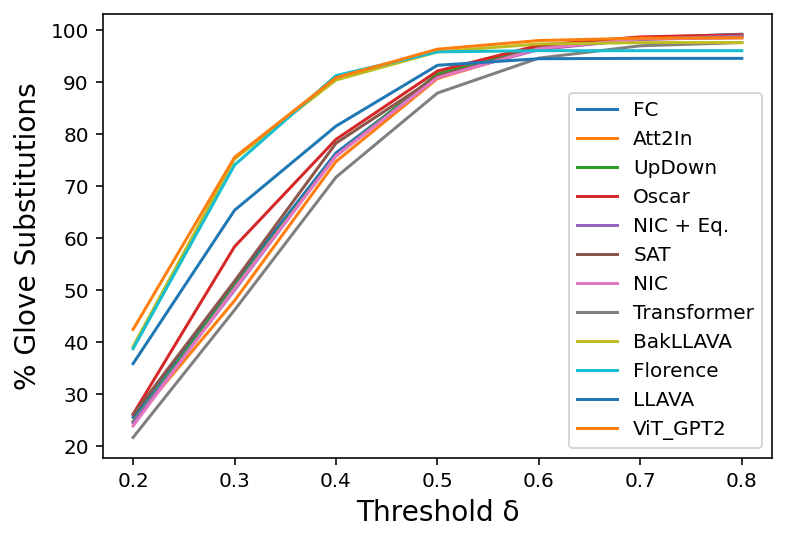}
    \caption{\textbf{Glove Thresholds:} Percentage of glove substitutions vs. the selected threshold $\delta$ across different models.}
    \label{fig: glove thresholds}
\end{figure}

To select the optimal distance threshold $\delta$ for our experiments, we observed how many contextual substitutions occurred at different thresholds. We tested this using the GloVe word embedding model. 

If $\delta$ is large, we allow weaker word substitutions. Consider the HGC: ``a $<$gender$>$ is sleeping''. Assume the word ``sleeping'' is not present in $V_{MGC}$. If $\delta$ is large, ``sleeping'' can get replaced with an unrelated word like ``book''. If $\delta$ is small, we only allow very similar word substitutions. But we increase the risk of $<$unk$>$ replacements. 

In Figure \ref{fig: glove thresholds}, the x-axis shows different distance thresholds ($\delta$), and the y-axis shows the $\%$ of contextual substitutions out of the total substitutions (contextual + $<$unk$>$). 

For GloVe, we found $\delta = 0.4$ to be the best balance. Below $0.4$, we had very less contextual substitutions ($< 60\%$). Above $\delta = 0.4$, we had a lot of contextual substitutions. However, a large number of contextual substitutions could be unrelated words (e.g., replacing sleeping with book).

\section{Experiment Setup: Additional Details}
\label{sec:additional_details}

For language-aware metrics (LIC and DBAC), we used an LSTM-unidirectional model as our sentence encoder (unless specified otherwise). We paired this LSTM with a 3-layer MLP to form our final attacker model.

For the controlled experiment \ref{subsec:lang_based_bias_exp}, we used accuracy as our quality function (Q). For experiments \ref{subsec:directionality}, \ref{subsec:sentence_encoder_sensitivity}, and \ref{subsec:constant_vs_contextual}, we set $Q = 1/\textit{cross-entropy}$.

We performed vocabulary alignment in our experiments using GloVe at $\delta = 0.4$ (unless specified otherwise). We found $\delta = 0.4$ to be a suitable threshold based on the sensitivity experiment we performed in Section \ref{sec: glove thresholds}. While FastText is better at vocabulary alignment, we used Glove to show that DBAC is superior even if we use a slightly weaker embedding model for alignment.

\section{Sensitivity to sentence encoders: Bias amplification scores}
\label{sec: Sensitivity Complete Table}

For gender, we reported DBAC$_{A \rightarrow T}$ scores on six sentence encoders trained from scratch on COCO captions, in Table \ref{tab:dbac_sensitivity_gender_scratch}. We reported the corresponding LIC scores in Table \ref{tab:lic_sensitivity_gender_scratch}. We reported DBAC$_{A \rightarrow T}$ scores on the four pre-trained encoders in Table \ref{tab:dbac_sensitivity_values_gender}. We reported the corresponding LIC scores in Table \ref{tab:lic_sensitivity_values_gender}. 

For race, we reported DBAC$_{A \rightarrow T}$ scores on six sentence encoders trained from scratch, in Table \ref{tab:DBAC_scratch_race_rnn}. We reported the corresponding LIC scores in Table \ref{tab:lic_scratch_race_rnn}. We reported DBAC$_{A \rightarrow T}$ scores on four pre-trained encoders in Table \ref{tab:dbac_sensitivity_values_race}. We reported the corresponding LIC scores in Table \ref{tab:lic_sensitivity_values_race}.

\section{Race results for constant vs. contextual substitution}
\label{sec: Race Experiments}

In column $2$ of Table \ref{tab:Meteor Constant vs. Context Race}, for race HGCs, we showed the METEOR similarity scores for constant substitution. Columns $3$ and $4$ show the METEOR scores for contextual substitution with Glove and FastText, respectively. The percentage values in green indicate the $\%$ increase in METEOR score achieved by contextual substitution, relative to constant substitution.

\textbf{For race, contextual substitution better retained the original meaning of the HGCs.} Across all captioning models, contextual substitution produced higher similarity scores than constant substitution. On average, contextual substitution improved the METEOR sentence similarity score by $1.6\%$ with Glove, and $2.2\%$ with FastText. 

In column $2$ of Table \ref{tab:DBAC Constant vs. Context Race}, we showed DBAC scores of the HGCs for race ($\omega_{H: A \rightarrow T}$) with constant substitution. Columns $3$ and $4$ show the DBAC scores for contextual substitution with Glove and FastText, respectively. We created an identical table for LIC scores in Table \ref{tab:LIC Constant vs. Context Race}.

\textbf{For race, DBAC and LIC reported higher bias in the HGCs with contextual substitution.} On average, contextual substitution increased DBAC scores by $5.2\%$ with Glove, and $4.4\%$ with FastText (Table \ref{tab:DBAC Constant vs. Context Race}). Contextual substitution increased LIC scores by $3.0\%$ with Glove, and $3.9\%$ with FastText (Table \ref{tab:LIC Constant vs. Context Race}).

\begin{table*}[htbp]
\centering
\footnotesize
\begin{tabularx}{0.9\textwidth}{YYYYYYY}
\toprule
\textbf{Models} & \textbf{LSTM} & \textbf{LSTM Bidirectional} & \textbf{RNN} & \textbf{RNN Bidirectional} & \textbf{Transformer (1 head)} & \textbf{Transformer (5 head)} \\
\midrule
Att2In      & $ \ \ \ 0.0590 $  & $ \ \ \ 0.0778 $  & $ \ \ \ 0.0540 $  & $ \ \ \ 0.0620 $  & $ \ \ \ 0.1109 $  & $ \ \ \ 0.2669 $ \\
BakLLAVA    & $ -0.0298 $  & $ -0.0637 $  & $ -0.0259 $  & $ -0.0396 $  & $ -0.0554 $  & $ -0.0848 $ \\
BLIP        & $ -0.0998 $  & $ -0.1102 $  & $ -0.1344 $  & $ -0.0545 $  & $ -0.1395 $  & $ -0.1442 $ \\
FC          & $ \ \ \ 0.0645 $  & $ \ \ \ 0.0836 $  & $ \ \ \ 0.0105 $  & $ \ \ \ 0.0775 $  & $ \ \ \ 0.1269 $  & $ \ \ \ 0.1822 $ \\
Florence    & $ -0.1025 $  & $ -0.1558 $  & $ -0.0600 $  & $ -0.1316 $  & $ -0.1060 $  & $ -0.1876 $ \\
LLAVA       & $ -0.0676 $  & $ -0.1005 $  & $ -0.0761 $  & $ -0.1298 $  & $ -0.0892 $  & $ -0.0855 $ \\
Oscar       & $ \ \ \ 0.1119 $  & $ \ \ \ 0.1407 $  & $ \ \ \ 0.1052 $  & $ \ \ \ 0.1013 $  & $ \ \ \ 0.1138 $  & $ \ \ \ 0.1641 $ \\
SAT         & $ \ \ \ 0.0761 $  & $ \ \ \ 0.0856 $  & $ \ \ \ 0.0830 $  & $ \ \ \ 0.0832 $  & $ \ \ \ 0.0717 $  & $ \ \ \ 0.1175 $ \\
NIC         & $ \ \ \ 0.2743 $  & $ \ \ \ 0.2467 $  & $ \ \ \ 0.1877 $  & $ \ \ \ 0.2694 $  & $ \ \ \ 0.1623 $  & $ \ \ \ 0.3122 $ \\
NIC+Equal   & $ \ \ \ 0.2552 $  & $ \ \ \ 0.2261 $  & $ \ \ \ 0.1417 $  & $ \ \ \ 0.2837 $  & $ \ \ \ 0.1843 $  & $ \ \ \ 0.3511 $ \\
Transformer & $ \ \ \ 0.1215 $  & $ \ \ \ 0.1358 $  & $ \ \ \ 0.1325 $  & $ \ \ \ 0.1511 $  & $ \ \ \ 0.1525 $  & $ \ \ \ 0.2335 $ \\
UpDn        & $ \ \ \ 0.1172 $  & $ \ \ \ 0.1193 $  & $ \ \ \ 0.1121 $  & $ \ \ \ 0.1091 $  & $ \ \ \ 0.1511 $  & $ \ \ \ 0.2357 $ \\
Vit\_GPT2   & $ -0.0206 $  & $ -0.0123 $  & $ -0.0841 $  & $ -0.0051 $  & $ -0.0279 $  & $ -0.0452 $ \\
\bottomrule
\end{tabularx}
\caption{DBAC scores for gender, on the six encoders trained from scratch on COCO captions.}
\label{tab:dbac_sensitivity_gender_scratch}
\end{table*}

\begin{table*}[htbp]
\centering
\footnotesize
\begin{tabularx}{0.9\textwidth}{YYYYYYY}
\toprule
\textbf{Models} & \textbf{LSTM} & \textbf{LSTM Bidirectional} & \textbf{RNN} & \textbf{RNN Bidirectional} & \textbf{Transformer (1 head)} & \textbf{Transformer (5 head)} \\
\midrule
Att2In      & $ \ \ \,0.0001 $ & $ \ \ \,0.0064 $ & $-0.0119$ & $-0.0294$ & $-0.0005$ & $ \ \ \,0.0004 $ \\
BakLLAVA    & $ \ \ \,0.0726 $ & $ \ \ \,0.0160 $ & $ \ \ \,0.0232 $ & $ \ \ \,0.0000 $ & $-0.0004$ & $-0.0019$ \\
BLIP        & $ \ \ \,0.0649 $ & $ \ \ \,0.0194 $ & $ \ \ \,0.0000 $ & $-0.0213$ & $-0.0005$ & $ \ \ \,0.0001 $ \\
FC          & $ \ \ \,0.0669 $ & $-0.0240$ & $-0.0225$ & $-0.0307$ & $-0.0009$ & $ \ \ \,0.0001 $ \\
Florence    & $-0.0236$ & $ \ \ \,0.0161 $ & $-0.0065$ & $-0.0068$ & $-0.0002$ & $-0.0005$ \\
LLAVA       & $ \ \ \,0.0021 $ & $ \ \ \,0.0383 $ & $-0.0002$ & $ \ \ \,0.0077 $ & $ \ \ \,0.0010 $ & $ \ \ \,0.0002 $ \\
Oscar       & $-0.0092$ & $ \ \ \,0.0385 $ & $-0.0154$ & $-0.0153$ & $ \ \ \,0.0003 $ & $-0.0026$ \\
SAT         & $ \ \ \,0.0405 $ & $ \ \ \,0.0162 $ & $ \ \ \,0.0099 $ & $ \ \ \,0.0159 $ & $-0.0001$ & $ \ \ \,0.0002 $ \\
NIC         & $ \ \ \,0.0107 $ & $ \ \ \,0.0242 $ & $ \ \ \,0.0337 $ & $-0.0018$ & $-0.0019$ & $-0.0006$ \\
NIC+Equal   & $ \ \ \,0.0408 $ & $ \ \ \,0.0009 $ & $-0.0462$ & $-0.0305$ & $ \ \ \,0.0005 $ & $ \ \ \,0.0003 $ \\
Transformer & $ \ \ \,0.0310 $ & $ \ \ \,0.0195 $ & $ \ \ \,0.0401 $ & $ \ \ \,0.0259 $ & $-0.0001$ & $ \ \ \,0.0000 $ \\
UpDn        & $ \ \ \,0.0549 $ & $ \ \ \,0.0000 $ & $-0.0114$ & $ \ \ \,0.0163 $ & $-0.0020$ & $ \ \ \,0.0002 $ \\
Vit\_GPT2   & $ \ \ \,0.0263 $ & $ \ \ \,0.0295 $ & $-0.0307$ & $ \ \ \,0.0000 $ & $-0.0015$ & $ \ \ \,0.0003 $ \\
\bottomrule
\end{tabularx}
\caption{LIC scores for gender, on the six encoders trained from scratch on COCO captions.}
\label{tab:lic_sensitivity_gender_scratch}
\end{table*}

\begin{table*}[htbp]
\centering
\footnotesize
\begin{subtable}[t]{0.47\textwidth}
\centering
\begin{tabularx}{\textwidth}{YYP{1.6cm}P{1.2cm}Y}
\toprule
\textbf{Models} & \textbf{DistilBERT} & \textbf{DistilRoBERTa} & \textbf{MiniLM} & \textbf{MPNet} \\
\midrule
Att2In       & $ \ \ \,0.0222 $  & $ \ \ \,0.0206 $  & $ \ \ \,0.0186 $  & $ \ \ \,0.0145 $ \\
BakLLAVA     & $ \ \ \,0.0680 $  & $ \ \ \,0.0774 $  & $ \ \ \,0.0804 $  & $ \ \ \,0.0475 $ \\
BLIP         & $-0.0044$  & $-0.0026$  & $-0.0004$  & $-0.0016$ \\
FC           & $-0.0231$  & $-0.0162$  & $-0.0016$  & $-0.0154$ \\
Florence     & $ \ \ \,0.1060 $  & $ \ \ \,0.1221 $  & $ \ \ \,0.1182 $  & $ \ \ \,0.0797 $ \\
LLAVA        & $ \ \ \,0.0979 $  & $ \ \ \,0.0740 $  & $ \ \ \,0.1038 $  & $ \ \ \,0.0581 $ \\
Oscar        & $ \ \ \,0.0442 $  & $ \ \ \,0.0368 $  & $ \ \ \,0.0457 $  & $ \ \ \,0.0328 $ \\
SAT          & $ \ \ \,0.0183 $  & $ \ \ \,0.0202 $  & $ \ \ \,0.0047 $  & $ \ \ \,0.0277 $ \\
NIC          & $ \ \ \,0.0850 $  & $ \ \ \,0.0166 $  & $ \ \ \,0.0185 $  & $ \ \ \,0.0103 $ \\
NIC+Equal    & $-0.0214$  & $-0.0040$  & $-0.0014$  & $-0.0081$ \\
Transformer  & $ \ \ \,0.0369 $  & $ \ \ \,0.0434 $  & $ \ \ \,0.0340 $  & $ \ \ \,0.0463 $ \\
UpDn         & $ \ \ \,0.0137 $  & $ \ \ \,0.0200 $  & $ \ \ \,0.0196 $  & $ \ \ \,0.0189 $ \\
Vit\_GPT2    & $ \ \ \,0.0154 $  & $ \ \ \,0.0175 $  & $ \ \ \,0.0334 $  & $ \ \ \,0.0216 $ \\
\bottomrule
\end{tabularx}
\caption{DBAC scores for gender on the pre-trained encoders}
\label{tab:dbac_sensitivity_values_gender}
\end{subtable}
\hfill
\begin{subtable}[t]{0.47\textwidth}
\centering
\begin{tabularx}{\textwidth}{YYP{1.6cm}P{1.2cm}Y}
\toprule
\textbf{Models} & \textbf{DistilBERT} & \textbf{DistilRoBERTa} & \textbf{MiniLM} & \textbf{MPNet} \\
\midrule
Att2In       & $-0.0013$  & $ \ \ \,0.0004 $  & $-0.0015$  & $-0.0014$ \\
BakLLAVA     & $ \ \ \,0.0013 $  & $-0.0005$  & $-0.0002$  & $-0.0001$ \\
BLIP         & $-0.0004$  & $-0.0001$  & $-0.0001$  & $-0.0005$ \\
FC           & $-0.0001$  & $-0.0002$  & $ \ \ \,0.0002 $  & $-0.0003$ \\
Florence     & $-0.0028$  & $-0.0001$  & $ \ \ \,0.0000 $  & $ \ \ \,0.0007 $ \\
LLAVA        & $-0.0032$  & $ \ \ \,0.0003 $  & $ \ \ \,0.0066 $  & $-0.0006$ \\
Oscar        & $ \ \ \,0.0018 $  & $ \ \ \,0.0003 $  & $-0.0010$  & $ \ \ \,0.0002 $ \\
SAT          & $-0.0008$  & $ \ \ \,0.0008 $  & $ \ \ \,0.0019 $  & $-0.0006$ \\
NIC          & $ \ \ \,0.0007 $  & $ \ \ \,0.0001 $  & $-0.0034$  & $ \ \ \,0.0001 $ \\
NIC+Equal    & $-0.0001$  & $-0.0006$  & $ \ \ \,0.0001 $  & $-0.0006$ \\
Transformer  & $ \ \ \,0.0000 $  & $ \ \ \,0.0000 $  & $ \ \ \,0.0011 $  & $-0.0006$ \\
UpDn         & $ \ \ \,0.0009 $  & $ \ \ \,0.0004 $  & $ \ \ \,0.0019 $  & $-0.0007$ \\
Vit\_GPT2    & $ \ \ \,0.0001 $  & $ \ \ \,0.0011 $  & $-0.0012$  & $-0.0008$ \\
\bottomrule
\end{tabularx}
\caption{LIC scores for gender on the pre-trained encoders}
\label{tab:lic_sensitivity_values_gender}
\end{subtable}
\caption{DBAC and LIC scores for gender, on the four pre-trained encoders}
\label{tab:dbac_lic_gender}
\end{table*}

\begin{table*}[htbp]
\centering
\footnotesize
\begin{tabularx}{0.9\textwidth}{YYYYYYY}
\toprule
\textbf{Models} & \textbf{LSTM} & \textbf{LSTM Bidirectional} & \textbf{RNN} & \textbf{RNN Bidirectional} & \textbf{Transformer (1 head)} & \textbf{Transformer (5 head)} \\
\midrule
Att2In      & $ \ \ \,0.1572 $ & $ \ \ \,0.2254 $ & $ \ \ \,0.1224 $ & $ \ \ \,0.1864 $ & $ \ \ \,0.1941 $ & $ \ \ \,0.1010 $ \\
BakLLAVA    & $-0.0945$ & $-0.0670$ & $-0.1804$ & $-0.1203$ & $-0.0810$ & $-0.0803$ \\
BLIP        & $ \ \ \,0.1557 $ & $ \ \ \,0.0446 $ & $ \ \ \,0.1125 $ & $ \ \ \,0.1683 $ & $ \ \ \,0.2367 $ & $ \ \ \,0.1894 $ \\
FC          & $ \ \ \,0.2754 $ & $ \ \ \,0.3091 $ & $ \ \ \,0.2207 $ & $ \ \ \,0.1572 $ & $ \ \ \,0.2689 $ & $ \ \ \,0.3115 $ \\
Florence    & $-0.4296$ & $-0.4052$ & $-0.4295$ & $-0.3960$ & $-0.5565$ & $-0.5210$ \\
LLAVA       & $-0.0676$ & $-0.1005$ & $-0.0761$ & $-0.1298$ & $-0.0892$ & $-0.0855$ \\
Oscar       & $ \ \ \,0.1975 $ & $ \ \ \,0.1655 $ & $ \ \ \,0.2962 $ & $ \ \ \,0.1461 $ & $ \ \ \,0.3409 $ & $ \ \ \,0.3939 $ \\
SAT         & $ \ \ \,0.1048 $ & $ \ \ \,0.1610 $ & $ \ \ \,0.1163 $ & $ \ \ \,0.1211 $ & $ \ \ \,0.1252 $ & $ \ \ \,0.0959 $ \\
NIC         & $ \ \ \,0.1104 $ & $ \ \ \,0.1140 $ & $ \ \ \,0.0835 $ & $ \ \ \,0.1753 $ & $ \ \ \,0.1961 $ & $ \ \ \,0.1677 $ \\
NIC+Equal   & $ \ \ \,0.2248 $ & $ \ \ \,0.2585 $ & $ \ \ \,0.3363 $ & $ \ \ \,0.2077 $ & $ \ \ \,0.3046 $ & $ \ \ \,0.3759 $ \\
Transformer & $ \ \ \,0.2083 $ & $ \ \ \,0.2046 $ & $ \ \ \,0.1894 $ & $ \ \ \,0.1905 $ & $ \ \ \,0.2861 $ & $ \ \ \,0.3129 $ \\
UpDn        & $ \ \ \,0.1223 $ & $ \ \ \,0.1668 $ & $ \ \ \,0.1477 $ & $ \ \ \,0.1480 $ & $ \ \ \,0.2165 $ & $ \ \ \,0.1611 $ \\
Vit\_GPT2   & $ \ \ \,0.2521 $ & $ \ \ \,0.1965 $ & $ \ \ \,0.3782 $ & $ \ \ \,0.1201 $ & $ \ \ \,0.3610 $ & $ \ \ \,0.3735 $ \\
\bottomrule
\end{tabularx}
\caption{DBAC scores for race, on the six encoders trained from scratch on COCO captions.}
\label{tab:DBAC_scratch_race_rnn}
\end{table*}

\begin{table*}[htbp]
\centering
\footnotesize
\begin{tabularx}{0.9\textwidth}{YYYYYYY}
\toprule
\textbf{Models} & \textbf{LSTM} & \textbf{LSTM Bidirectional} & \textbf{RNN} & \textbf{RNN Bidirectional} & \textbf{Transformer (1 head)} & \textbf{Transformer (5 head)} \\
\midrule
Att2In      & $ \ \ \,0.0133 $ & $ \ \ \,0.0127 $ & $ \ \ \,0.0111 $ & $-0.0332$ & $ \ \ \,0.0298 $ & $ \ \ \,0.0629 $ \\
BakLLAVA    & $ \ \ \,0.0293 $ & $ \ \ \,0.0272 $ & $ \ \ \,0.0232 $ & $-0.0555$ & $ \ \ \,0.0304 $ & $ \ \ \,0.0373 $ \\
BLIP        & $ \ \ \,0.0111 $ & $ \ \ \,0.0046 $ & $ \ \ \,0.0112 $ & $ \ \ \,0.0213 $ & $ \ \ \,0.0066 $ & $ \ \ \,0.0036 $ \\
FC          & $-0.0404$ & $-0.0206$ & $-0.0444$ & $-0.0444$ & $-0.0263$ & $-0.0200$ \\
Florence    & $ \ \ \,0.0379 $ & $ \ \ \,0.0364 $ & $ \ \ \,0.0333 $ & $ \ \ \,0.0254 $ & $ \ \ \,0.0458 $ & $ \ \ \,0.0734 $ \\
LLAVA       & $ \ \ \,0.0221 $ & $ \ \ \,0.0413 $ & $ \ \ \,0.0333 $ & $ \ \ \,0.0077 $ & $ \ \ \,0.0472 $ & $ \ \ \,0.0502 $ \\
Oscar       & $ \ \ \,0.0200 $ & $ \ \ \,0.0373 $ & $ \ \ \,0.0333 $ & $ \ \ \,0.0167 $ & $ \ \ \,0.0430 $ & $ \ \ \,0.0618 $ \\
SAT         & $ \ \ \,0.0419 $ & $ \ \ \,0.0391 $ & $ \ \ \,0.0333 $ & $ \ \ \,0.0111 $ & $ \ \ \,0.0400 $ & $ \ \ \,0.0581 $ \\
NIC         & $ \ \ \,0.0036 $ & $ \ \ \,0.0032 $ & $ \ \ \,0.0000 $ & $ \ \ \,0.0111 $ & $ \ \ \,0.0097 $ & $ \ \ \,0.0245 $ \\
NIC+Equal   & $ \ \ \,0.0176 $ & $ \ \ \,0.0167 $ & $ \ \ \,0.0111 $ & $-0.0665$ & $ \ \ \,0.0157 $ & $ \ \ \,0.0342 $ \\
Transformer & $ \ \ \,0.0105 $ & $ \ \ \,0.0084 $ & $ \ \ \,0.0000 $ & $-0.0444$ & $ \ \ \,0.0146 $ & $ \ \ \,0.0340 $ \\
UpDn        & $ \ \ \,0.0106 $ & $ \ \ \,0.0092 $ & $ \ \ \,0.0000 $ & $ \ \ \,0.0111 $ & $ \ \ \,0.0108 $ & $ \ \ \,0.0114 $ \\
Vit\_GPT2   & $-0.0104$ & $-0.0265$ & $-0.0222$ & $-0.0212$ & $-0.0080$ & $-0.0062$ \\
\bottomrule
\end{tabularx}
\caption{LIC scores for race, on the six encoders trained from scratch on COCO captions.}
\label{tab:lic_scratch_race_rnn}
\end{table*}

\begin{table*}[htbp]
\centering
\footnotesize
\begin{subtable}[t]{0.47\textwidth}
\centering
\begin{tabularx}{\textwidth}{YYP{1.6cm}P{1.2cm}Y}
\toprule
\textbf{Models} & \textbf{DistilBERT} & \textbf{DistilRoBERTa} & \textbf{MiniLM} & \textbf{MPNet} \\
\midrule
Att2In      & $ \ \ \,0.0432 $ & $ \ \ \,0.0626 $ & $ \ \ \,0.0587 $ & $ \ \ \,0.0519 $ \\
BakLLAVA    & $-0.1794$ & $-0.1696$ & $-0.1460$ & $-0.1715$ \\
BLIP        & $-0.0084$ & $-0.0093$ & $-0.0268$ & $-0.0116$ \\
FC          & $ \ \ \,0.0162 $ & $ \ \ \,0.0218 $ & $ \ \ \,0.0297 $ & $ \ \ \,0.0238 $ \\
Florence    & $-0.5405$ & $-0.5185$ & $-0.4997$ & $-0.5155$ \\
LLAVA       & $-0.1507$ & $-0.1789$ & $-0.1632$ & $-0.1805$ \\
Oscar       & $ \ \ \,0.0366 $ & $ \ \ \,0.0127 $ & $ \ \ \,0.0107 $ & $ \ \ \,0.0320 $ \\
SAT         & $ \ \ \,0.0108 $ & $ \ \ \,0.0121 $ & $ \ \ \,0.0260 $ & $ \ \ \,0.0232 $ \\
NIC         & $ \ \ \,0.0337 $ & $ \ \ \,0.0103 $ & $ \ \ \,0.0273 $ & $ \ \ \,0.0231 $ \\
NIC+Equal   & $ \ \ \,0.0109 $ & $ \ \ \,0.0044 $ & $ \ \ \,0.0230 $ & $ \ \ \,0.0223 $ \\
Transformer & $ \ \ \,0.0205 $ & $ \ \ \,0.0127 $ & $ \ \ \,0.0265 $ & $ \ \ \,0.0182 $ \\
UpDn        & $ \ \ \,0.0214 $ & $ \ \ \,0.0159 $ & $ \ \ \,0.0287 $ & $ \ \ \,0.0329 $ \\
Vit\_GPT2   & $ \ \ \,0.0433 $ & $ \ \ \,0.0374 $ & $ \ \ \,0.0380 $ & $ \ \ \,0.0166 $ \\
\bottomrule
\end{tabularx}
\caption{DBAC scores for race on the pre-trained encoders}
\label{tab:dbac_sensitivity_values_race}
\end{subtable}
\hfill
\begin{subtable}[t]{0.47\textwidth}
\centering
\begin{tabularx}{\textwidth}{YYP{1.6cm}P{1.2cm}Y}
\toprule
\textbf{Models} & \textbf{DistilBERT} & \textbf{DistilRoBERTa} & \textbf{MiniLM} & \textbf{MPNet} \\
\midrule
Att2In      & $ \ \ \,0.0011 $ & $ \ \ \,0.0002 $ & $ \ \ \,0.0004 $ & $ \ \ \,0.0001 $ \\
BakLLAVA    & $ \ \ \,0.0003 $ & $-0.0005$ & $-0.0002$ & $-0.0003$ \\
BLIP        & $ \ \ \,0.0011 $ & $ \ \ \,0.0002 $ & $ \ \ \,0.0003 $ & $-0.0001$ \\
FC          & $ \ \ \,0.0003 $ & $-0.0001$ & $ \ \ \,0.0005 $ & $-0.0003$ \\
Florence    & $ \ \ \,0.0005 $ & $ \ \ \,0.0004 $ & $ \ \ \,0.0014 $ & $ \ \ \,0.0000 $ \\
LLAVA       & $ \ \ \,0.0049 $ & $-0.0018$ & $ \ \ \,0.0007 $ & $-0.0031$ \\
Oscar       & $ \ \ \,0.0012 $ & $ \ \ \,0.0000 $ & $ \ \ \,0.0000 $ & $ \ \ \,0.0004 $ \\
SAT         & $-0.0012$ & $-0.0004$ & $ \ \ \,0.0003 $ & $-0.0001$ \\
NIC         & $ \ \ \,0.0003 $ & $-0.0003$ & $ \ \ \,0.0004 $ & $-0.0001$ \\
NIC+Equal   & $ \ \ \,0.0002 $ & $-0.0002$ & $ \ \ \,0.0002 $ & $-0.0002$ \\
Transformer & $ \ \ \,0.0023 $ & $ \ \ \,0.0000 $ & $-0.0001$ & $ \ \ \,0.0000 $ \\
UpDn        & $ \ \ \,0.0006 $ & $-0.0002$ & $ \ \ \,0.0008 $ & $ \ \ \,0.0005 $ \\
Vit\_GPT2   & $ \ \ \,0.0002 $ & $ \ \ \,0.0001 $ & $ \ \ \,0.0002 $ & $-0.0001$ \\
\bottomrule
\end{tabularx}
\caption{LIC scores for race on the pre-trained encoders}
\label{tab:lic_sensitivity_values_race}
\end{subtable}
\caption{DBAC and LIC scores for race, on the four pre-trained encoders.}
\label{tab:dbac_lic_race}
\end{table*}

\begin{table*}[htbp]
\centering
\footnotesize
\begin{tabularx}{0.47\textwidth}{P{1.2cm}P{1cm}P{2.2cm}P{2.2cm}}
\toprule
\textbf{Models} & \textbf{Constant} & \multicolumn{2}{c}{\textbf{Contextual}} \\
\cmidrule(lr){3-4}
 & & \textbf{Glove} & \textbf{FastText} \\
\midrule
Att2In      & $0.709$ & $0.732$ (\green{$3.22$\% $\uparrow$}) & $0.741$ (\green{$4.44$\% $\uparrow$})\\
BakLLAVA    & $0.901$ & $0.901$ (\green{$0.01$\% $\uparrow$}) & $0.901$ (\green{$0.01$\% $\uparrow$})\\
BLIP        & $0.855$ & $0.856$ (\green{$0.08$\% $\uparrow$}) & $0.855$ (\green{$0.01$\% $\uparrow$})\\
FC          & $0.654$ & $0.680$ (\green{$3.90$\% $\uparrow$}) & $0.686$ (\green{$4.94$\% $\uparrow$})\\
Florence    & $0.881$ & $0.881$ (\green{$0.00$\% $\uparrow$}) & $0.881$ (\green{$0.00$\% $\uparrow$})\\
LLAVA       & $0.891$ & $0.891$ (\green{$0.01$\% $\uparrow$}) & $0.891$ (\green{$0.00$\% $\uparrow$})\\
Oscar       & $0.706$ & $0.724$ (\green{$2.66$\% $\uparrow$}) & $0.731$ (\green{$3.56$\% $\uparrow$})\\
SAT         & $0.790$ & $0.793$ (\green{$0.38$\% $\uparrow$}) & $0.805$ (\green{$1.86$\% $\uparrow$})\\
NIC         & $0.804$ & $0.805$ (\green{$0.19$\% $\uparrow$}) & $0.806$ (\green{$0.24$\% $\uparrow$})\\
NIC+Equal.  & $0.772$ & $0.794$ (\green{$2.84$\% $\uparrow$}) & $0.805$ (\green{$4.22$\% $\uparrow$})\\
Transformer & $0.837$ & $0.850$ (\green{$1.56$\% $\uparrow$}) & $0.853$ (\green{$1.91$\% $\uparrow$})\\
UpDown      & $0.735$ & $0.764$ (\green{$3.90$\% $\uparrow$}) & $0.769$ (\green{$4.63$\% $\uparrow$})\\
VIT\_GPT2   & $0.815$ & $0.829$ (\green{$1.66$\% $\uparrow$}) & $0.839$ (\green{$2.92$\% $\uparrow$})\\
\midrule
\textbf{Average} & $0.796$ & $0.808$ (\green{$1.57$\% $\uparrow$}) & $0.812$ (\green{$2.21$\% $\uparrow$})\\
\bottomrule
\end{tabularx}
\caption{METEOR scores for constant vs. contextual substitution (race results): Values in green indicate $\%$ increase in METEOR scores with contextual substitution, relative to constant substitution. For all captioning models, contextual substitution achieved a higher METEOR score than constant substitution.}
\label{tab:Meteor Constant vs. Context Race}
\end{table*}

\begin{table*}[htbp]
\centering
\footnotesize
\begin{subtable}[t]{0.45\textwidth}
\centering
\begin{tabularx}{\textwidth}{P{1cm}P{0.9cm}P{2cm}P{2.2cm}}
\toprule
\textbf{Models} & \textbf{Constant} & \multicolumn{2}{c}{\textbf{Contextual}} \\
\cmidrule(lr){3-4}
 & & \textbf{Glove} & \textbf{FastText} \\
\midrule
Att2In      & $0.290$ & $0.300$ (\green{$3.49$\% $\uparrow$}) & $0.309$ (\green{$6.45$\% $\uparrow$})\\
BakLLAVA    & $0.281$ & $0.283$ (\green{$0.93$\% $\uparrow$}) & $0.284$ (\green{$1.10$\% $\uparrow$})\\
BLIP        & $0.287$ & $0.291$ (\green{$1.11$\% $\uparrow$}) & $0.291$ (\green{$1.29$\% $\uparrow$})\\
FC          & $0.284$ & $0.291$ (\green{$2.68$\% $\uparrow$}) & $0.286$ (\green{$0.74$\% $\uparrow$})\\
Florence    & $0.281$ & $0.307$ (\green{$9.52$\% $\uparrow$}) & \ $0.314$ (\green{$12.09$\%$\uparrow$})\\
LLAVA       & $0.274$ & $0.278$ (\green{$1.57$\% $\uparrow$}) & $0.275$ (\green{$0.37$\% $\uparrow$})\\
Oscar       & \ $0.305$ & $0.342$ (\green{$12.38$\%$\uparrow$}) & $0.329$ (\green{$7.88$\% $\uparrow$})\\
SAT         & $0.294$ & $0.299$ (\green{$1.84$\% $\uparrow$}) & $0.307$ (\green{$4.50$\% $\uparrow$})\\
NIC         & $0.265$ & $0.290$ (\green{$9.44$\% $\uparrow$}) & $0.269$ (\green{$1.70$\% $\uparrow$})\\
NIC+Equal.  & $0.300$ & $0.302$ (\green{$0.93$\% $\uparrow$}) & $0.302$ (\green{$0.67$\% $\uparrow$})\\
Transformer & $0.263$ & \ $0.291$ (\green{$10.76$\%$\uparrow$}) & $0.280$ (\green{$6.46$\% $\uparrow$})\\
UpDown      & $0.273$ & $0.294$ (\green{$7.70$\% $\uparrow$}) & $0.288$ (\green{$5.42$\% $\uparrow$})\\
VIT\_GPT2   & $0.300$ & $0.316$ (\green{$5.36$\% $\uparrow$}) & $0.326$ (\green{$8.73$\% $\uparrow$})\\
\midrule
\textbf{Average} & $0.284$ & $0.299$ (\green{$5.21$\% $\uparrow$}) & $0.297$ (\green{$4.42$\% $\uparrow$})\\
\bottomrule
\end{tabularx}
\caption{DBAC race results for the HGCs: $\omega_{H:A \rightarrow T}$}
\label{tab:DBAC Constant vs. Context Race}
\end{subtable}
\hfill
\begin{subtable}[t]{0.45\textwidth}
\centering
\begin{tabularx}{\textwidth}{P{1cm}P{0.9cm}P{2cm}P{2cm}}
\toprule
\textbf{Models} & \textbf{Constant} & \multicolumn{2}{c}{\textbf{Contextual}} \\
\cmidrule(lr){3-4}
 & & \textbf{Glove} & \textbf{FastText} \\
\midrule
Att2In      & $0.518$ & $0.539$ (\green{$4.07$\% $\uparrow$}) & $0.545$ (\green{$5.15$\% $\uparrow$})\\
BakLLAVA    & $0.521$ & $0.528$ (\green{$1.27$\% $\uparrow$}) & $0.539$ (\green{$3.47$\% $\uparrow$})\\
BLIP        & $0.581$ & $0.609$ (\green{$4.70$\% $\uparrow$}) & $0.591$ (\green{$1.67$\% $\uparrow$})\\
FC          & $0.522$ & $0.540$ (\green{$3.33$\% $\uparrow$}) & $0.530$ (\green{$1.44$\% $\uparrow$})\\
Florence    & $0.517$ & $0.532$ (\green{$2.86$\% $\uparrow$}) & $0.532$ (\green{$2.84$\% $\uparrow$})\\
LLAVA       & $0.544$ & $0.555$ (\green{$1.93$\% $\uparrow$}) & $0.548$ (\green{$0.74$\% $\uparrow$})\\
Oscar       & $0.522$ & $0.536$ (\green{$2.70$\% $\uparrow$}) & $0.545$ (\green{$4.41$\% $\uparrow$})\\
SAT         & $0.536$ & $0.545$ (\green{$1.64$\% $\uparrow$}) & $0.558$ (\green{$4.18$\% $\uparrow$})\\
NIC         & $0.545$ & $0.563$ (\green{$3.30$\% $\uparrow$}) & $0.580$ (\green{$6.41$\% $\uparrow$})\\
NIC+Equal.  & $0.545$ & $0.563$ (\green{$3.26$\% $\uparrow$}) & $0.562$ (\green{$3.15$\% $\uparrow$})\\
Transformer & $0.518$ & $0.527$ (\green{$1.72$\% $\uparrow$}) & $0.528$ (\green{$1.97$\% $\uparrow$})\\
UpDown      & $0.519$ & $0.534$ (\green{$2.87$\% $\uparrow$}) & $0.547$ (\green{$5.37$\% $\uparrow$})\\
VIT\_GPT2   & $0.496$ & $0.521$ (\green{$5.17$\% $\uparrow$}) & \ $0.548$ (\green{$10.47$\%$\uparrow$})\\
\midrule
\textbf{Average} & $0.530$ & $0.545$ (\green{$2.99$\% $\uparrow$}) & $0.550$ (\green{$3.94$\% $\uparrow$})\\
\bottomrule
\end{tabularx}
\caption{LIC race results for the HGCs}
\label{tab:LIC Constant vs. Context Race}
\end{subtable}
\vspace{-0.3cm}
\caption{DBAC and LIC scores for constant vs. contextual substitution (race results): Values in green indicate the $\%$ increase in HGC's bias reported by contextual substitution relative to constant substitution. Across all captioning models, both DBAC and LIC consistently reported higher bias in the HGCs with contextual substitution.}
\vspace{-0.3cm}
\label{tab:DBAC and LIC Constant vs. Contextual Race}
\end{table*}

\end{document}